%% file: main.tex
\documentclass[10pt,twocolumn,letterpaper]{article}

\usepackage[pagenumbers]{cvpr}      % To produce the REVIEW version
\usepackage{pifont}
\usepackage{multirow}
\usepackage{comment}
\usepackage{rotating}
\usepackage{makecell}
\usepackage{tabularx}
\usepackage{float}
\input{preamble}

\definecolor{cvprblue}{rgb}{0.21,0.49,0.74}
\usepackage[pagebackref,breaklinks,colorlinks,citecolor=cvprblue]{hyperref}

\def\paperID{2644} % *** Enter the Paper ID here
\def\confName{CVPR}
\def\confYear{2024}

\title{SEED-Bench-2: Benchmarking Multimodal Large Language Models}

\author{
Bohao Li$^{3,1\star}$ \and
Yuying Ge$^{1\star}$ \and
Yixiao Ge$^{1,2\dagger}$ \and
Guangzhi Wang$^{2}$ \and
Rui Wang$^{1}$ \and
Ruimao Zhang$^{3\dagger}$ \and
Ying Shan$^{1,2}$ \and
\\
$^{1}$Tencent AI Lab \\
$^{2}$ARC Lab, Tencent PCG \\
$^{3}$School of Data Science, The Chinese University of HongKong, Shenzhen \\
}

\begin{document}

\twocolumn[{%
\renewcommand\twocolumn[1][]{#1}%
    \maketitle
    \begin{figure}[H]
        \hsize=\textwidth
        \centering
        \begin{subfigure}{0.6\textwidth}
            \centering
            \includegraphics[width=\linewidth]{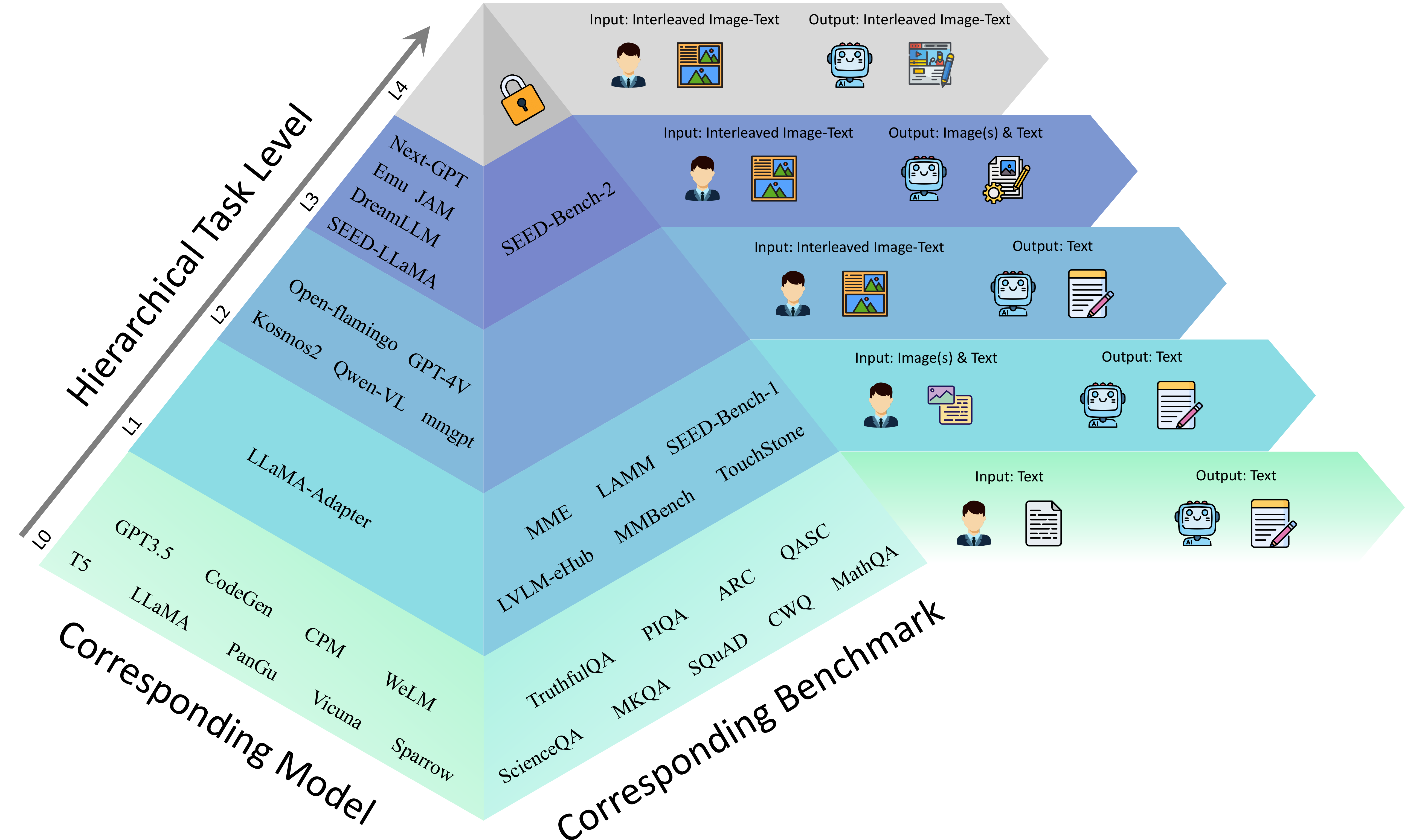}
        \end{subfigure}
        \hfill
        \begin{subfigure}{0.39\textwidth}
            \centering
            \includegraphics[width=\linewidth]{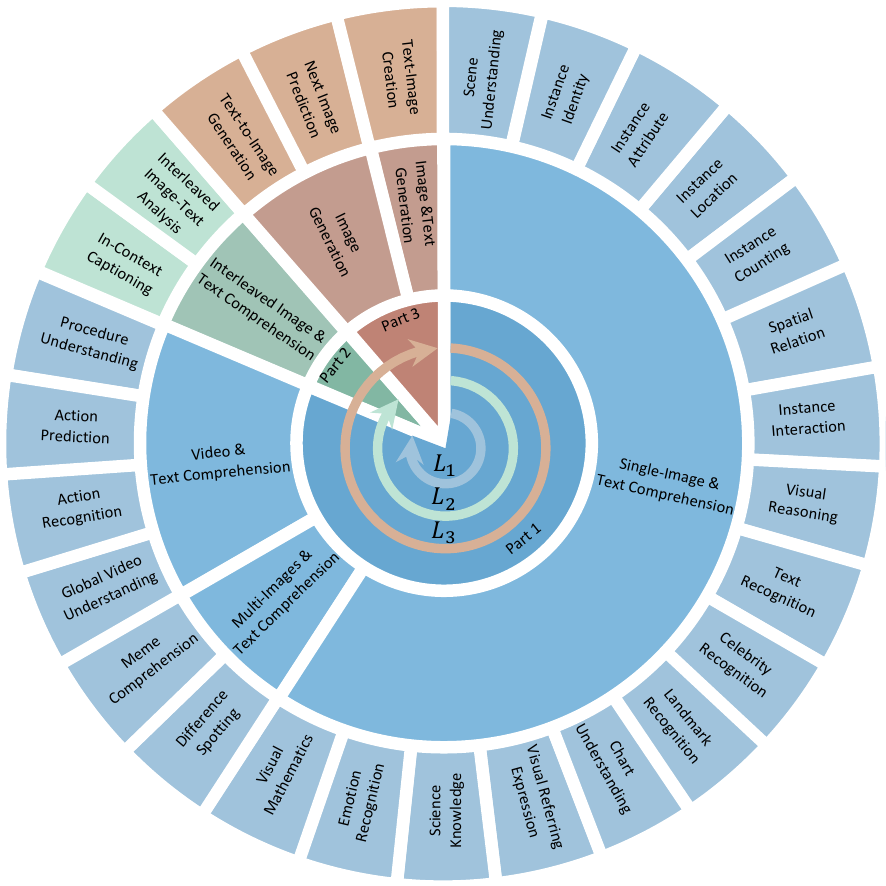}
        \end{subfigure}
        \hfill
        \vspace{5pt}
        \caption{(left) Overview of \textbf{hierarchical capability levels} of MLLMs from $L_0$ to $L_4$, where higher level encompasses lower capability tiers. Models and corresponding evaluation benchmarks at each pyramid tier are illustrated. SEED-Bench-2 covers the assessment of MLLMs up to $L_3$. (right) Overview of 27 evaluation dimensions in SEED-Bench-2, which consists of three parts, with part-1 constituting $L_1$, part-1\&2 constituting $L_2$, and part-1\&2\&3 constituting $L_3$.} 
        %(right) The leaderboard of different MLLMs in SEED-Bench-2 (part-1\&2\&3).}
        %models and benchmarks across various task levels. The bars at different levels of the pyramid represent the data formats of tasks at each level, while the models and benchmarks at each level indicate the capabilities encompassed and assessed by the corresponding models and benchmarks. Higher-level models and benchmarks include the content and capabilities of lower-level models and benchmarks.}
        \label{fig:teaser}
    \end{figure}
}]

\renewcommand{\thefootnote}{\fnsymbol{footnote}}
 		\footnotetext[1]{Equal Contribution.} 
   \footnotetext[2]{Correspondence Author.}

\input{sec/0_abstract}    
\input{sec/1_intro}
\input{sec/2_related}
\input{sec/3_seed_bench}
\input{sec/4_results}
\input{sec/5_conclusion}
{
    \small
    \bibliographystyle{ieeenat_fullname}
    \bibliography{main}
}

\input{sec/X_suppl}

\end{document}

%% file: preamble.tex
\usepackage[dvipsnames]{xcolor}

%% file: sec/0_abstract.tex
\begin{table*}[]
    \centering
    \caption{Comparisons between existing MLLM benchmarks. ``H/G Evaluation'' denotes whether human or GPT is used for evaluation.}\label{tab:benchmark_compare}
    % \vspace{-0.6em}
    \vspace{-3pt}
    {\small
    \resizebox{\textwidth}{!}{
    \begin{tabular}{cccccccc}
         \toprule
             \multirow{1}{*}{Benchmark} & \multirow{1}{*}{Visual Modality} & \multirow{1}{*}{Evaluation Level} & \multirow{1}{*}{Customized Question} & \multirow{1}{*}{\#Answer Annotation}& \multirow{1}{*}{Answer Type} & \multirow{1}{*}{H/G Evaluation} & \multirow{1}{*}{\#Models} \\
         \midrule
         LLaVA-Bench~\cite{liu2023visual_llava} & Image &$L_1$ & \ding{51} & 150 & free-form & GPT & 4\\
         OCR-Bench~\cite{liu2023ocrbench} & Image & $L_1$ & \ding{55} & - &free-form & N/A & 6 \\
         MME~\cite{fu2023mme} & Image  &$L_1$ & \ding{51} & 2194 &Y/N & N/A & 10\\
         M3Exam~\cite{zhang2023m3exam} & Image &$L_1$ &\ding{51} & 12317 & A/B/C/D & N/A & 7\\
         LAMM~\cite{yin2023lamm} & Image(s) \& Point cloud &$L_1$ & \ding{55} & - & free-form & GPT & 4\\
         LVLM-eHub~\cite{xu2023lvlm} & Image & $L_1$  & \ding{55} & - & free-form & Human &8\\
         MMBench~\cite{liu2023mmbench} & Image(s) &$L_1$  &\ding{51} &2974 & free-form & GPT &14\\
         VisIT-Bench~\cite{bitton2023visitbench} &Image{s} & $L_1$ & \ding{51} & 592 &free-form &Human/GPT &14\\
         MM-VET~\cite{yu2023mmvet} & Image &$L_1$ &\ding{51} & 200 &free-form & GPT &9 \\
         Touchstone~\cite{bai2023touchstone} & Image(s) &$L_1$&\ding{51} & 908 & free-form & GPT & 7\\
         SciGraphQA~\cite{li2023scigraphqa} & Image & $L_1$ &\ding{51} & 3K &free-form &N/A & 4 \\
         SEED-Bench-1~\cite{li2023seed} & Image(s) \& Video & $L_1$ &\ding{51} & 19242 & A/B/C/D & N/A & 18\\
         SEED-Bench-2 & Image(s) \& Video &$L_3$& \ding{51} & 24371 & A/B/C/D & N/A & 23\\
         \bottomrule
    \vspace{-20pt}
    \end{tabular}
    }
   }
\end{table*}

\begin{abstract}
Multimodal large language models (MLLMs), building upon the foundation of powerful large language models (LLMs), have recently demonstrated exceptional capabilities in generating not only texts but also images given interleaved multimodal inputs (acting like a combination of GPT-4V and DALL-E 3). 
However, existing MLLM benchmarks remain limited to assessing only models' comprehension ability of single image-text inputs, failing to keep up with the strides made in MLLMs. A comprehensive benchmark is imperative for investigating the progress and uncovering the limitations of current MLLMs. In this work, we categorize the capabilities of MLLMs into hierarchical levels from $L_0$ to $L_4$ based on the modalities they can accept and generate, and propose SEED-Bench-2, a comprehensive benchmark that evaluates the \textbf{hierarchical} capabilities of MLLMs. Specifically, SEED-Bench-2 comprises 24K multiple-choice questions with accurate human annotations, which spans 27 dimensions, including the evaluation of both text and image generation. Multiple-choice questions with groundtruth options derived from human annotation enables an objective and efficient assessment of model performance, eliminating the need for human or GPT intervention during evaluation. We further evaluate the performance of 23 prominent open-source MLLMs and summarize valuable observations. By revealing the limitations of existing MLLMs through extensive evaluations, we aim for SEED-Bench-2 to provide insights that will motivate future research towards the goal of General Artificial Intelligence.
Dataset and evaluation code are available at \href{https://github.com/AILab-CVC/SEED-Bench}{\color{magenta}https://github.com/AILab-CVC/SEED-Bench.}
\end{abstract}

%% file: sec/1_intro.tex
\section{Introduction}

In recent years, Large Language Models (LLMs)~\cite{chung2022scaling_flant5, openai2023gpt4, ChatGPT, vicuna, touvron2023llama} have exhibited remarkable capabilities to understand, reason, and generate texts across a variety of open-ended tasks. Leveraging the strong generality of LLMs, Multimodal Large Language Models (MLLMs)~\cite{li2023blip2, zhu2023minigpt4, liu2023visual_llava, ye2023mplugowl, dai2023instructblip, li2023otter, gong2023multimodalgpt, su2023pandagpt, peng2023kosmos, li2023videochat, maaz2023videochatgpt, luo2023valley, peng2023kosmos, bai2023qwen, liu2023llava1.5, laurencon2023obelics, zhang2023internlm} have demonstrated exceptional capabilities in comprehending multimodal data through predicting open-form texts. Recent work~\cite{sun2023emu, yu2023scaling, ge2023planting, ge2023making, wu2023nextgpt, dong2023dreamllm} further empower LLMs with the ability of generating images beyond texts (acting like a combination of GPT-4V~\cite{2023GPT4VisionSC} and DALL-E 3~\cite{BetkerImprovingIG}), since they contend that the premise for the emergence of multimodal capabilities is that text and image can be represented and processed interchangeably in a unified autoregressive Transformer. However, despite the extensive capabilities of MLLMs, existing MLLM benchmarks~\cite{fu2023mme,yin2023lamm,xu2023lvlm,liu2023mmbench,bai2023touchstone} primarily focus on evaluating single image-text comprehension, thus failing to fully demonstrate the progress and limitations of current MLLMs. The lag of benchmarks behind the rapid development of MLLMs hinders the exploration and evolution of models.

In this work, we categorize the capabilities of MLLMs into hierarchical levels ranging from $L_0$ to $L_4$ based on the modalities they can accept and generate, as depicted in Fig.~\ref{fig:teaser}. Building upon LLMs, the lowest-tier capability $L_0$ involves generating texts given text inputs, while the highest-tier capability $L_4$ entails producing open-form interleaved image and text output given arbitrary interleaved image-text inputs. Reaching the capability $L_4$ is a crucial milestone on the path towards General Artificial Intelligence (AGI) since a human-level AI should be able to effortlessly digest and create multimodal content. In the capability pyramid, higher levels inherently include the capabilities of lower tiers. This hierarchical categorization not only clearly illustrates the current progress of MLLMs, but also provides a well-defined roadmap for future research.

We propose SEED-Bench-2\footnote{This benchmark inherits the evaluation dimensions from SEED-Bench-1~\cite{li2023seed}, which constitutes a part of capability level $L_1$.}, a comprehensive benchmark that evaluates the \textbf{hierarchical} capabilities of MLLMs up to $L_3$, including the generation of both texts and images given interleaved image-text inputs. As shown in Fig.~\ref{fig:teaser}, SEED-Bench-2 consists of three parts, where part-1 constitutes capability level $L_1$ for images and texts comprehension, part-1\&2 constitute capability level $L_2$ for interleaved image-text comprehension, and part-1\&2\&3 constitute capability level $L_3$ for image and text generation. To the best of our knowledge, SEED-Bench-2 is the first benchmark that provides hierarchical evaluations of MLLMs, which effectively showcases the range of model capabilities.

%evaluates the generation of both texts and images given interleaved image-text inputs. It consists of three parts, where part-1 examines the comprehension of fixed-form images and texts as shown in Fig.~\ref{fig:part1}, part-2 assesses the understanding of open-form interleaved image-text inputs, and part-3 evaluates the generation of both texts and images as shown in Fig.~\ref{fig:part23}. Part-1 constitutes the assessment benchmark of $L_1$ while part-2 together with part-1 forms the benchmark of $L_2$. To the best of our knowledge, SEED-Bench-2 is the first benchmark that provides hierarchical evaluations of MLLMs, which clearly illustrates the scope of model capabilities.

%
Specifically, SEED-Bench-2 consists of 24K multiple-choice questions with groundtruth answers derived from human annotation (×10 larger than MME~\cite{fu2023mme} and ×8 larger than MMBench~\cite{liu2023mmbench} as shown in Tab.~\ref{tab:benchmark_compare}). SEED-Bench-2 spans 27 evaluation dimensions, enabling a comprehensive assessment of MLLMs' performance across diverse aspects. We employ three approaches for the generation of multiple-choice questions, including (1) a sophisticated pipeline utilizing foundation models,  (2) the adaptation of existing datasets, and (3) a combination of human creation and GPT assistance. We further incorporate automated filtering mechanism and manual verification process to ensure the quality of questions and the accuracy of groundtruth answers. Different from existing MLLM benchmarks~\cite{liu2023visual_llava, xu2023lvlm, yin2023lamm, bai2023touchstone, liu2023mmbench, yu2023mmvet, bitton2023visitbench} that employ human annotators or GPT to evaluate open-form output, resulting in compromised efficiency, increased subjectivity, and reduced assessment accuracy, SEED-Bench-2 provides multiple-choice questions, which restricts the model’s output to A/B/C/D options. This approach facilitates the convenient computation of accuracy, serving as an an objective metric for evaluation.

Based on SEED-Bench-2, we comprehensively evaluate 23 prominent open-source MLLMs. Our evaluation results yield the following three key findings: (1) Existing MLLMs have not yet reached the ceiling level of capability $L_1$ for the comprehension of fixed-form images and texts, with even the top-ranked model achieving only a 60\% accuracy rate. MLLMs, in particular, exhibit poor performance in certain dimensions, such as understanding charts and visual mathematics. (2) MLLMs achieve less satisfactory performance at capability $L_2$ than that at $L_1$, which indicates that it is more challenging for MLLMs to comprehend free-form interleaved image-text inputs, since most MLLMs are trained on structured image-caption pairs. (3) At present, only a few MLLMs can attain capability $L_3$, which requires models to output content in multiple modalities. A universal MLLM that unifies the generation of images and texts is currently underexplored. We will launch an evaluation platform and consistently maintain a leaderboard for assessing and comparing model performance.

\begin{figure*}
    % \centering
    \includegraphics[width=1.0\textwidth]{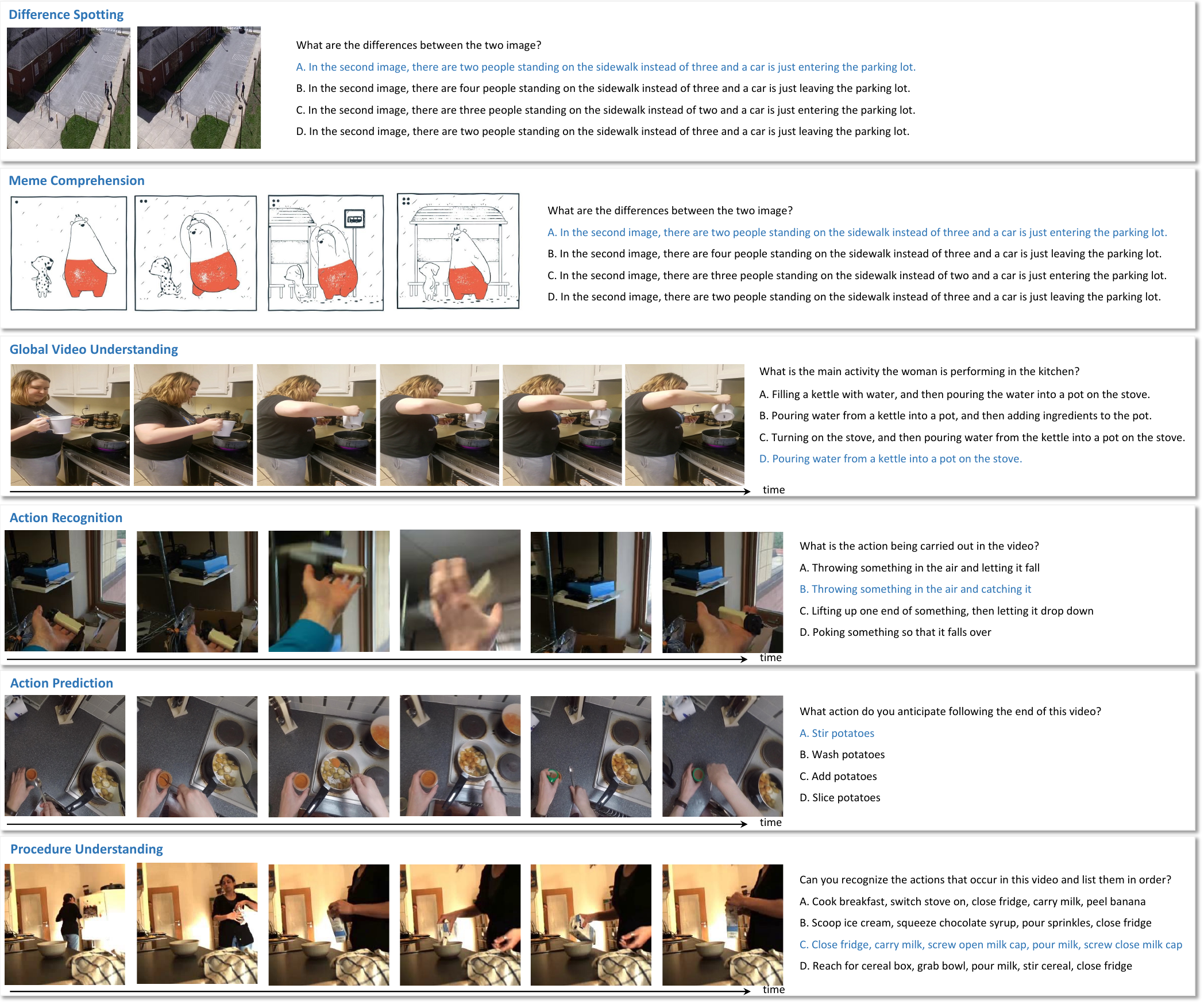}
    \caption{Data samples from a subset of evaluation dimensions in part-1 with multiple images or videos as inputs, which encompasses capability $L_1$ in SEED-Bench-2.}
    \label{fig:part1}
        \vspace{-12pt}
\end{figure*}

\begin{figure*}
    % \centering
    \includegraphics[width=1\textwidth]{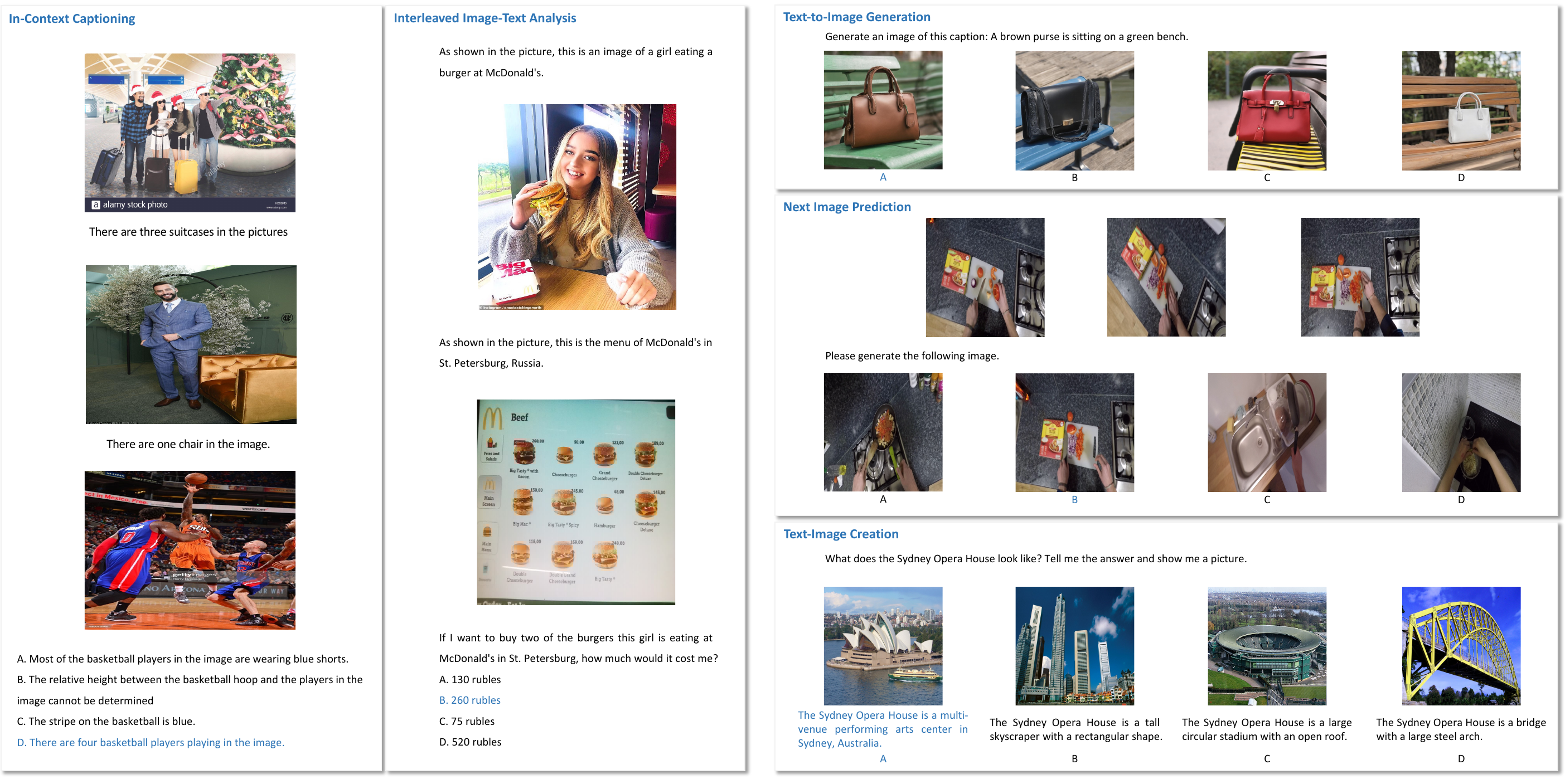}
    \caption{(left) Data samples of evaluation dimensions in part-2 with interleaved image-text as inputs, which encompasses capability $L_2$ together with dimensions in $L_1$. (right) Data samples of evaluation dimensions in part-3 with images and texts as outputs, which encompasses capability $L_3$ together with dimensions in $L_2$.}
    \label{fig:part23}
        \vspace{-12pt}
\end{figure*}

%% file: sec/2_related.tex
\section{Related Work}\label{sec:related_work}
\noindent\textbf{Multimodal Large Language Models.}
With the impressive success of Large language models (LLM)~\cite{chung2022scaling_flant5,touvron2023llama,vicuna}, recent studies work on generative Multimodal Large Language Models (MLLMs)~\cite{li2023blip2, zhu2023minigpt4, liu2023visual_llava, ye2023mplugowl, dai2023instructblip, li2023otter, gong2023multimodalgpt, su2023pandagpt,peng2023kosmos, bai2023qwen, liu2023llava1.5, laurencon2023obelics, zhang2023internlm} to improve multimodal comprehension through aligning visual features of pre-trained image encoder with LLMs on image-text datasets. Some work~\cite{li2023videochat,maaz2023videochatgpt,luo2023valley} further considers video inputs and leverage the vast capabilities of LLMs for video understanding tasks. Recent work~\cite{sun2023emu, yu2023scaling, ge2023planting, ge2023making, wu2023nextgpt, dong2023dreamllm} take significant strides in equipping MLLMs with the capacity for generating images beyond texts. In SEED-Bench-2, we provide a comprehensive and objective evaluation of these models to thoroughly assess their hierarchical capabilities. 

\noindent\textbf{Benchmarks for Multimodal Large Language Models.}
With the rapid development of Multimodal Large Language Models (MLLMs), some concurrent works~\cite{fu2023mme,yin2023lamm,xu2023lvlm,liu2023mmbench,bai2023touchstone} propose various benchmarks for evaluating MLLMs. However, they remain limited to assessing only model's ability of predicting texts given single image-text inputs, failing to keep up with the strides made in multimodal model capabilities. 
For example, GVT~\cite{wang2023gvt} constructs a benchmark by aggregating two semantic-level understanding tasks (VQA and Image Captioning) and two fine-grained tasks (Object Counting and Multi-class Identification). But its evaluation is constrained to limited aspects of visual understanding. LVLM-eHub~\cite{xu2023lvlm} combines multiple existing computer vision benchmarks and develops an online platform, where two models are prompted to answer a question related to an image and human annotators are employed to compare the predictions of models. The involvement of human annotators during evaluation not only introduces bias but also incurs significant costs. LLaVA-Bench~\cite{liu2023visual_llava}, LAMM~\cite{yin2023lamm} and Touchstone~\cite{bai2023touchstone} utilize GPT to evaluate the answers’ relevance and accuracy to the groundtruth. The reliance on entity extraction and GPT metric can impact the accuracy and reliability of the evaluation. MME~\cite{fu2023mme} and MMBench~\cite{liu2023mmbench} aim to enhance the objective evaluation of MLLMs by constructing 2194 True/False Questions and 2974 Multiple Choice Questions across a variety of ability dimensions respectively. %introduce a benchmark of 27 dimensions consisting of 908 questions to evaluate MLLMs. 
Considering the limited scale of these benchmarks, their evaluation results may exhibit instability. In this work, we introduce SEED-Bench-2 to evaluates the hierarchical capabilities of MLLMs including the generation of both texts and images, which contains 24K human-annotated multiple-choice questions covering 27 evaluation dimensions.

%% file: sec/3_seed_bench.tex
\section{SEED-Bench-2}

\subsection{Hierarchical Capability Levels}
We categorize the capabilities of MLLMs into hierarchical levels from $L_0$ to $L4$, based on input and output modalities, where
higher level encompasses lower capability tiers, as illustrated in Fig.~\ref{fig:teaser}. SEED-Bench-2 covers the assessment of MLLMs up to $L_3$. The detailed categorization of capability level is illustrated as below,

{\flushleft \bf \textbf{Level $L_0$}}: Building upon LLMs, the most fundamental capability of MLLMs generating text based on provided text inputs, which does not necessitate specific evaluation within the MLLM benchmark.
{\flushleft \bf \textbf{Level $L_1$}}: MLLMs at this capability level should possess the ability to comprehend multimodal inputs in a fixed format, \textit{i.e.}, image or multiple images (video input can be regarded as multiple images) and then texts. Current MLLM benchmarks only evaluate this capability level with single image and text as inputs.
{\flushleft \bf \textbf{Level $L_2$}}: MLLMs at this capability level should be able to understand multimodal inputs with open-form interleaved image-text data, which aligns with the multimodal inputs encountered in real-life scenarios.
{\flushleft \bf \textbf{Level $L_3$}}: Besides the inherent ability of LLMs to generate texts, MLLMs at this capability level should also be proficient in producing images, as advanced MLLMs are expected to process and represent multimodal content on both input and output sides.
{\flushleft \bf \textbf{Level $L_4$}}: MLLMs at the highest capability level should possess the ability to process and generate interleaved image-text content in an open-form format, which is an essential step towards achieving general artificial intelligence. We will incorporate evaluations of this capability level in our future work.

\subsection{Evaluation Dimensions}
As shown in Fig.~\ref{fig:teaser}, SEED-Bench-2 comprises a total of 27 evaluation dimensions, which constitute three capabilities level, from $L_1$ to $L_3$. Since higher level encompasses lower capability tiers, we further divide the evaluation dimensions of $L_3$ to three non-overlapping parts: part-1 forms level $L_1$, part-2 combined with part-1 constitute level $L_2$, part-3, part-2 and part-1 form level $L_3$ together. We introduce the dimensions of each part in details as below,

\subsubsection{Part-1}
The dimensions of part-1 evaluate MLLMs' comprehension of multimodal inputs in a fixed format, and can be further grouped into three sub-parts based on the types of visual inputs: (1) Single-Image \& Text, (2) Multiple-Images \& Text, (3) Video \& Text.

\begin{itemize}
\item Single-Image \& Text Comprehension. This sub-part consists of diverse evaluation dimensions including Scene Understanding, Instance Identity, Instance Attribute, Instance Location, Instance Counting, Spatial Relation, Instance Interaction, Visual Reasoning, Text Recognition, Celebrity Recognition, Landmark Recognition, Chart Understanding, Visual Referring Expression, Science Knowledge, Emotion Recognition and Visual Mathematics. These dimensions assess MLLMs' comprehension of image-text pair from extensive aspects, encompassing global/object-level understanding, recognition/reasoning, and various specialized domains.

\item Multiple-Images \& Text Comprehension. This sub-part contains Difference Spotting and Meme Comprehension, which evaluates MLLMs' capability of extracting information and discerning differences given multiple images.

\item Video \& Text Comprehension. This sub-part consists of Global Video Understanding, Action Recognition, Action Prediction and Procedure Understanding, which assesses MLLMs' ability for fine-grained action recognition, temporal relationship understanding and temporal reasoning.

\end{itemize}

\subsubsection{Part-2}
Part-2 evaluate MLLMs' comprehension of arbitrary interleaved image-text inputs, including In-Context Captioning, where two examples of image-caption pairs and an image are given, and the model is expected to describe the specific aspect of the image, and Interleaved Image-Text Analysis, where the model answers questions based on images and texts with varying quantities and positions.

\subsubsection{Part-3}
The dimensions of part-3 evaluate MLLMs' capability of generating images in addition to texts, and can be divided into two sub-parts including (1) Image generation and (2) Image \& Text generation.

\begin{itemize}
\item Image generation. This sub-part comprises Text-to-Image Generation, where the model is expected to generate an image based on a caption prompt, and Next Image Generation, where the model is required to generate an subsequent image based on previous images.

\item Text-Image creation. Given a question, the model is required to provide a text-based answer and subsequently generate a corresponding image as an illustration.
\end{itemize}

\begin{figure*}
    % \centering
    \includegraphics[width=1\textwidth]%#{figs/SeedBench_pipeline_fig_latex.pdf}
    {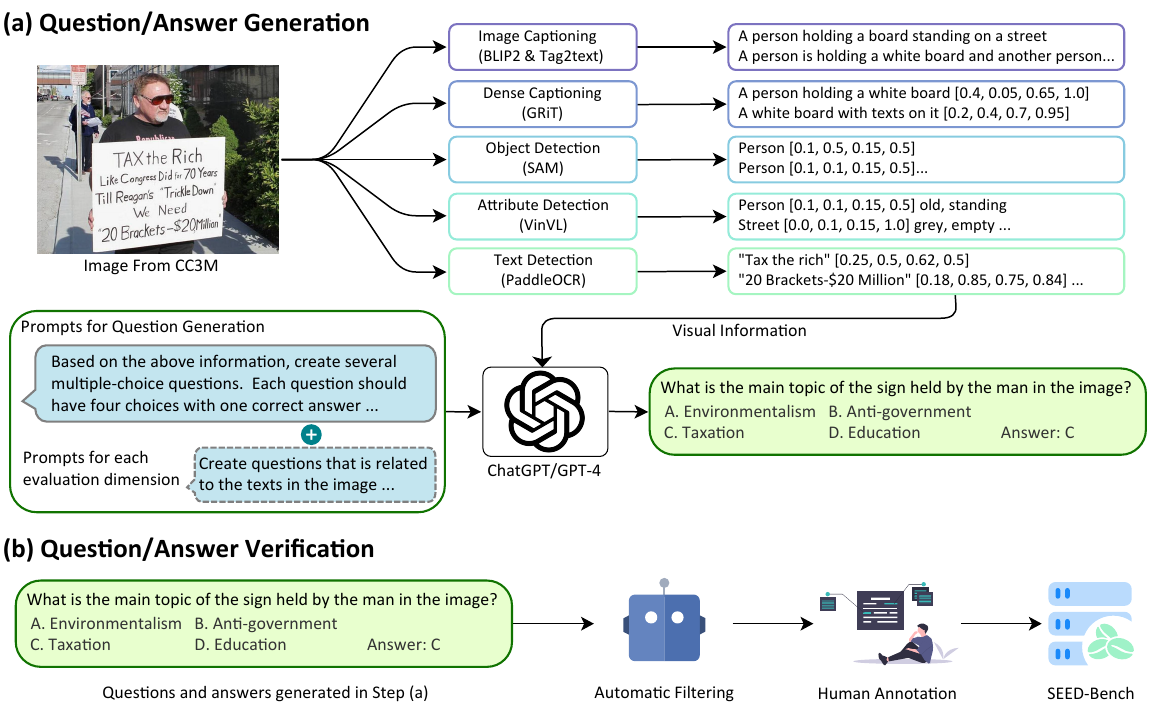}
    \caption{Overview of automatic pipeline in SEED-Bench-2 for generating multiple-choice questions. (a) We first leverage various foundation models to extract visual information including image-level captions, instance-level descriptions and textual elements. Based on specially designed prompts corresponding to specific evaluation dimension, ChatGPT/GPT-4 subsequently  generates questions and four candidate options with one groundtruth answer. (b) We further filter out questions by utilizing LLMs and employ human annotators to select the correct option and classify each question into one evaluation dimension.}
    \label{fig_pipeline}
    \vspace{-12pt}
\end{figure*}

\subsection{Construction of Multiple-choice Questions}
We employ three approaches to construct multiple-choice question covering 27 evaluation dimensions: (1) an automatic pipeline to generate questions for specific evaluation dimension, (2) tailor of existing datasets for the format of multiple-choice questions, (3) human creation combined with GPT. The details of the construction of each evaluation dimension can be found in the supplementary material.

\paragraph{Automatic pipeline.}
As shown in Fig.~\ref{fig_pipeline}, our pipeline for generating multiple-choice questions involves question/answer generation and verification. For generating question/answer pairs, we first leverage various foundation models to extract visual information including image-level captions, instance-level descriptions and textual elements. Based on specially designed prompts corresponding to specific evaluation dimension, ChatGPT/GPT-4 subsequently generates questions and four candidate options with one groundtruth answer. For verifying question/answer pairs, we filter out questions that can be answered correctly by multiple LLMs without resorting to visual information, since such questions are not helpful to evaluate the visual comprehension capability of MLLMs. We further employ human annotators to select the correct option and classify each question into one evaluation dimension. 

\paragraph{Tailoring existing datasets.} For existing datasets with annotated label, we first prompt ChatGPT/GPT-4 to generate questions based on provided information. We then construct distracting choices either from the annotated labels of other samples or by utilizing ChatGPT to generate three distractors. For distractors generated by ChatGPT, we additionally utilize human annotators to filter out options that are too similar to the groundtruth answer.

\paragraph{Human creation combined with GPT.} For evaluation dimensions lacking suitable data, \textit{e.g. Interleaved Image-Text Analysis} and \textit{Text-Image Creation}, we employ human annotators to meticulously design questions, retrieve corresponding images, and construct distracting choices with the assistance of ChatGPT.

\begin{table*}[]
    \centering
    \caption{Evaluation results of various MLLMs in different capability levels of SEED-Bench-2. $\bar{T}$ denotes the averaged accuracy across corresponding dimensions, and $R_{\bar{T}}$ denotes the rank based on the the averaged accuracy. The evaluation dimensions of part-2, together with $L_1$, encompass $L_2$, while the evaluation dimensions of part-3, together with $L_2$, encompass $L_3$.}\label{tab:performance}
    \vspace{3pt}
    % \vspace{-0.6em}
    {\small
    \resizebox{\textwidth}{!}{
    \begin{tabular}{cccccccccccc}
         \toprule
         \multirow{2}{*}{Model} & \multirow{2}{*}{Language Model}& \multicolumn{2}{c}{$L_1$ (Part-1)} & \multicolumn{2}{c}{Part-2} & \multicolumn{2}{c}{$L_2$} & \multicolumn{2}{c}{Part-3} & \multicolumn{2}{c}{$L_3$}\\
         \cmidrule(lr){3-4}
         \cmidrule(lr){5-6}
         \cmidrule(lr){7-8}
         \cmidrule(lr){9-10}
         \cmidrule(lr){11-12}
         && $\bar{T}$ & $R_{\bar{T}}$ & $\bar{T}$ & $R_{\bar{T}}$ & $\bar{T}$ & $R_{\bar{T}}$ & $\bar{T}$ & $R_{\bar{T}}$ & $\bar{T}$ & $R_{\bar{T}}$\\
         \midrule
         % \multirow{3}{*}{LLM} & Flan-T5~\cite{chung2022scaling_flant5} &Flan-T5-XL &27.6 & 23 &27.8 & 25 &\bf{50.9} &1 &\bf{46.7} & 1 & - & - & - & -\\
         % & Vicuna~\cite{vicuna} &Vicuna-7B &27.5 &24 &27.8 &24 &46.7 &3 &40.2 &5 & - & - & - & -\\
         % & LLaMA~\cite{touvron2023llama} &LLaMA-7B &27.0 &25 &28.3 &23 &43.2 &5 &38.9 &6 & - & - & - & -\\
         % \midrule
         BLIP-2~\cite{li2023blip2} &Flan-T5-XL &41.0 &9 &35.3 &10 & 40.5 & 8  & - & - & - & - \\
         InstructBLIP~\cite{dai2023instructblip} &Flan-T5-XL  &42.2 &7 &35.7 &6 & 41.7 & 7  & - & - & - & -\\
         InstructBLIP Vicuna~\cite{dai2023instructblip} &Vicuna-7B  &41.4 &8 &29.7 &19 & 40.5 & 9  & - & - & - & -\\
         LLaVA~\cite{liu2023visual_llava} &LLaMA-7B &38.7 &12 &30.2 &18 & 38.0 & 13  & - & - & - & -\\
         MiniGPT-4~\cite{zhu2023minigpt4} &Vicuna-7B &39.4 &10 &34.1 &13 & 39.0 & 10 & - & - & - & -\\
         VPGTrans~\cite{2023vpgtrans} &LLaMA-7B &36.2 &20 &23.9 &21 & 35.2 & 19  & - & - & - & -\\
         MultiModal-GPT~\cite{gong2023multimodalgpt} &Vicuna-7B &37.4 &15 &34.9 &12 & 37.1 & 14 & - & - & - & -\\
         Otter~\cite{li2023otter} &LLaMA-7B &36.4 &18 &36.6 &5 &36.4 & 17  & - & - & - & -\\
         OpenFlamingo~\cite{openflamingo} &LLaMA-7B &37.3 &16 &35.5 &9 & 37.1 & 15  & - & - & - & -\\
         LLaMA-Adapter V2~\cite{gao2023llamaadapterv2} &LLaMA-7B &37.5 &14 &- &- &- &-  & - & - & - & - \\
         GVT~\cite{wang2023gvt} &Vicuna-7B &34.4 &22 &38.6 &4 & 34.8 & 20  & - & - & - & -\\
         mPLUG-Owl~\cite{ye2023mplugowl} &LLaMA-7B &39.4 &11 &28.9 &20 & 38.5 & 11  & - & - & - & -\\ 
         Kosmos-2~\cite{peng2023kosmos} & Decoder only 1.3B &46.3 &3 &23.3 &22 &44.4 &3 & - & - & - & -\\
         Qwen-VL-Chat~\cite{bai2023qwen} & Qwen-7B &43.1 &5 &35.5 &8 & 42.5 & 5  & - & - & - & -\\
         LLaVA-1.5~\cite{liu2023llava1.5} &Vicuna-7B  &47.3 &2 &30.8 &17 & 46.0 & 2  & - & - & - & -\\
         IDEFICS-9B-Instruct~\cite{laurencon2023obelics} & LLaMA-7B &38.0 &13 &40.3 &3 & 38.2 & 12  & - & - & - & -\\
         InternLM-Xcomposer-VL~\cite{zhang2023internlm} & InternLM-7B &\bf{59.2} &1 &32.1 &15 & \bf{56.9} & 1  & - & - & - & -\\
         VideoChat~\cite{li2023videochat} &Vicuna-7B &37.0 &17 &35.3 &10 & 36.8 & 16  & - & - & - & -\\
         Video-ChatGPT~\cite{maaz2023videochatgpt} &LLaMA-7B &36.4 &19 &31.0 &16 & 35.9 & 18  & - & - & - & -\\
         Valley~\cite{luo2023valley} &LLaMA-13B &34.5 &21 &32.2 &14 & 34.3 & 21 & - & - & - & -\\
         Emu~\cite{sun2023emu} & LLaMA-13B &42.5 &6 &{41.1} &2 & 42.4 & 6 & {41.4} & 2 & {42.3} & 2\\
         NExt-GPT~\cite{wu2023nextgpt} &Vicuna-7B &30.7 &23  &35.6 &7 & 31.1 & 22 & 33.9 & 3 & 31.4 & 3\\
         SEED-LLaMA~\cite{ge2023making} & LLaMA2-Chat-13B &43.9 &4 &\bf{43.4} &1 &43.8 &4 & \bf{52.3} & 1 &\bf{44.8} &1\\
         \bottomrule
    \end{tabular}
    }
   }
\end{table*}
\subsection{Evaluation Strategy}
\label{sec:strategy}
\paragraph{Evaluation of text output.}
Different from MMBench~\cite{liu2023mmbench} that employs ChatGPT to match a model’s prediction to one of the choices in a multiple-choice question (achieves only 87.0\% alignment rate), we adopt the answer ranking strategy~\cite{dai2023instructblip, brown2020gpt3, lin2021truthfulqa} for evaluating existing MLLMs with multiple-choice questions. Specifically, for each choice of a question, we compute the likelihood that an MLLM generates the content of this choice given the question. We select the choice with the highest likelihood as model's prediction. Our evaluation strategy does not rely on the instruction-following capabilities of models to output ``A'' or ``B'' or ``C'' or ``D''. Furthermore, this evaluation strategy eliminates the impact of the order of multiple-choice options on the model's performance. 
\paragraph{Evaluation of image output.} Since not all MLLMs with image generation capabilities employ visual autoregression, adopting an answer ranking strategy for image evaluation is impractical. Instead, we calculate the CLIP similarity score~\cite{radford2021clip} between the generated image and each candidate image option, selecting the the highest-scoring option as the final prediction of the given multiple-choice question.
\paragraph{Evaluation of text and image output.} We first employ an answer ranking strategy to select the most likely text prediction. If it matches the ground truth, we evaluate the image output using the CLIP similarity score~\cite{radford2021clip} between the generated image and each candidate. The model is deemed correct only if both text and image predictions match the ground truth.

%% file: sec/4_results.tex
\section{Evaluation Results}
\subsection{Models}
\begin{figure*}
    % \centering
    \includegraphics[width=1.0\textwidth]{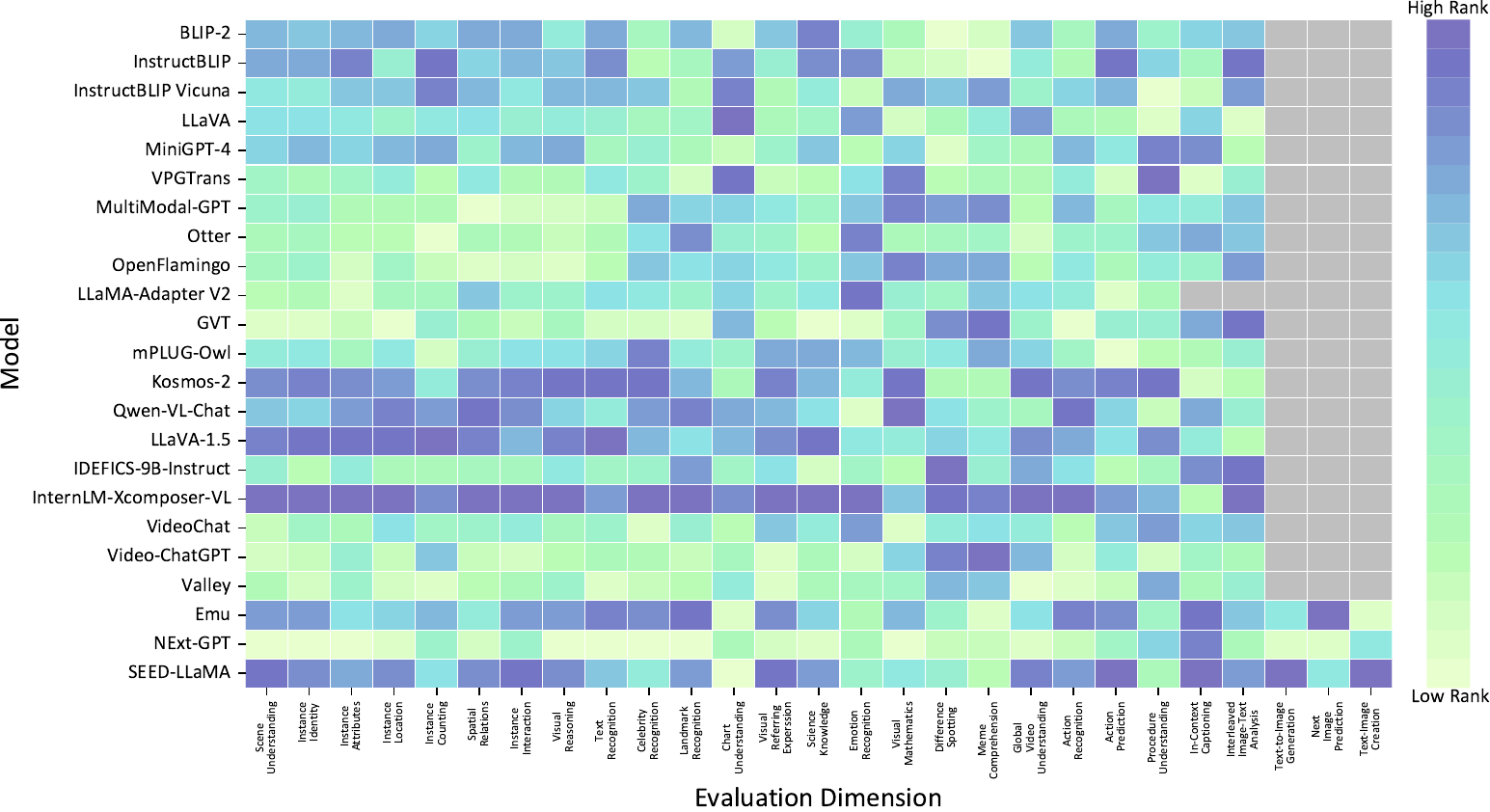}
    \caption{Illustration of each model's performance across different evaluation dimensions, where darker colors represent higher ranks. Gray indicates that the model has not yet reached the capability level required for evaluating that dimension.}
    \label{fig:rank}
    \vspace{-10pt}
\end{figure*}
We evaluate a total of 23 open-source MLLMs including BLIP-2~\cite{li2023blip2}, InstructBLIP~\cite{dai2023instructblip}, InstructBLIP Vicuna~\cite{dai2023instructblip}, LLaVA~\cite{liu2023visual_llava}, MiniGPT-4~\cite{zhu2023minigpt4}, VPGTrans~\cite{2023vpgtrans},  MultiModal-GPT~\cite{gong2023multimodalgpt}, Otter~\cite{li2023otter}, OpenFlamingo~\cite{openflamingo}, LLaMA-Adapter V2~\cite{gao2023llamaadapterv2}, GVT~\cite{wang2023gvt}, mPLUG-Owl~\cite{ye2023mplugowl}, Kosmos-2~\cite{peng2023kosmos}, Qwen-VL-Chat~\cite{bai2023qwen}, LLaVA1.5~\cite{liu2023llava1.5}, IDEFICS-9B-Instruct~\cite{laurencon2023obelics}, InternLM-Xcomposer-VL~\cite{zhang2023internlm}, VideoChat~\cite{li2023videochat}, Video-ChatGPT~\cite{maaz2023videochatgpt}, Valley~\cite{luo2023valley}, Emu~\cite{sun2023emu}, NExt-GPT~\cite{wu2023nextgpt}, and SEED-LLaMA~\cite{ge2023making} based on their official implementations. For each model, we first determine its capability level and then evaluate the corresponding dimensions. Note that we have confirmed with the authors that the LLaMA-Adapter V2's capability level is $L_1$. Some MLLMs can reach the capability level $L_3$, but they are not available as open-source.

\subsection{Main Results}
The evaluation results of various MLLMs in different capability levels of SEED-Bench-2 are listed in Tab.~\ref{tab:performance}. The detailed leaderboard of each evaluation dimension are provided in the supplemental materials. InternLM-Xcomposer-VL outperformes a large number of MLLMs, achieving the best performance based on the averaged accuracy in capability level $L_1$ and $L_2$, and Emu ranks top-1 in capability level $L_3$ with only one competitor. Because InternLM-Xcomposer-VL retrieves images from the available image pool rather than generate images, it does not reach the capability level $L_3$.  
To better showcase the the capabilities of models across different evaluation dimensions, we further visualize the ranking of each model within each evaluation dimension in Fig.~\ref{fig:rank}, where darker colors represent higher ranks and grey color indicates that the model has not yet reached the capability level required for evaluating that dimension. The champion MLLM InternLM-Xcomposer-VL achieves competitive results in a large number of evaluation dimensions of capability level $L_1$ and $L_2$. Although NExt-GPT reaches the capability level $L_3$, it performs poolry in multiple evaluation dimensions at level $L_1$ and $L_2$.

\subsection{Observations}
Through the comprehension and objective evaluation of various MLLMs in different capability levels of SEED-Bench-2, we have uncovered insights that can inform future work.

\noindent\textbf{Existing MLLMs have yet to reach the ceiling level of capability $L_1$.} Even the top-ranked MLLM achieves only a 60\% averaged accuracy in capability $L_1$, which evaluates the comprehension of multimodal inputs in a fixed format, \textit{i.e.}, images or multiple images (videos) and then texts. 

\noindent\textbf{The comprehension of Interleaved Image-Text data is more difficult.} 
The majority of MLLMs achieve worse results on part 2, which consists of multiple-choice questions with interleaved image-text inputs, than that on $L_1$ with fixed-form image and text as inputs.

\noindent\textbf{Only a small number of MLLMs can reach the capability $L_3$.} Only three open-source MLLMs possess the ability to generate images, besides the inherent ability of LLMs to output texts. A universal MLLM that unifies the generation of images and texts is currently underexplored. 

\noindent\textbf{It is challenging to address multimodal comprehension and generation simultaneously.} Although NExt-GPT reaches the capability level $L_3$, which can generate both texts and images, it shows poor performance in capability $L_1$ for multimodal comprehension. Equipping MLLMs with image generation ability without compromising their inherent text output performance remains to be addressed.

\noindent\textbf{All MLLMs struggle with understanding charts and visual mathematics.} The top-performing MLLMs achieves only around 30\% accuracy, which indicates that the understanding capabilities of MLLMs within specialized domains need enhancement.

\noindent\textbf{MLLMs trained on Interleaved Image-Text data excel in similar-format questions.} SEED-LLaMA, Emu, IDEFICS-9B-Instruct and Otter achieve higher accuracy in part 2, which consists of multiple-choice questions with interleaved image-text inputs. These MLLMs are trained on interleaved image-text data besides structured image-caption pairs, which demonstrates the importance of data for MLLM training.

\noindent\textbf{VideoLLMs fail to achieve competitive performance on temporal understanding.} 
Despite being instruction-tuned on video data, Video-ChatGPT and Valley underperform in temporal understanding compared to MLLMs pre-trained on image data. It indicates that current VideoLLMs have limited capabilities for fine-grained action recognition and temporal reasoning.

%% file: sec/5_conclusion.tex
\section{Conclusion}
In this work, we introduce SEED-Bench-2, a large-scale benchmark for evaluating Multimodal Large Language Models (MLLMs) in terms of hierarchical capabilities, including the generation of both texts and images. SEED-Bench-2 consists of 24K multiple-choice questions with accurate human annotations, which covers 27 evaluation dimensions. We conduct a thorough evaluation of 22 prominent open-source MLLMs, analyzing and comparing their performances to provide insights for future research. We plan to launch and maintain a leaderboard, offering a platform for the community to assess model performance. 

%% file: sec/X_suppl.tex
\clearpage
\maketitlesupplementary
\begin{figure}
    \centering
    \begin{subfigure}{0.5\textwidth}
        \centering
        \includegraphics[width=\linewidth]{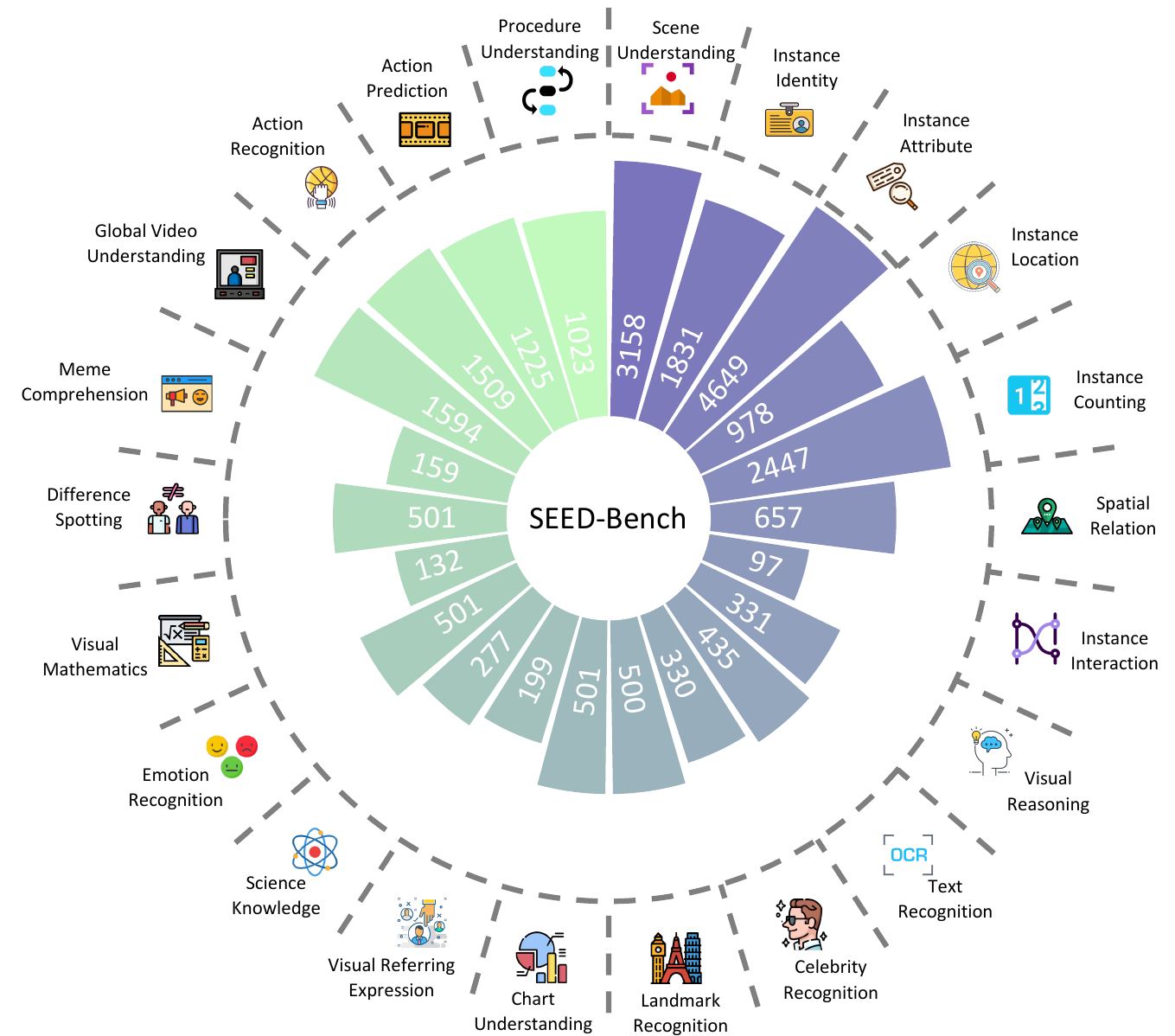}
    \end{subfigure}
    \caption{Overview of 22 evaluation dimensions in SEED-Bench-2 capability $L_1$. The number in the bar denotes the number of multiple-choice questions in each dimension.}
    \label{fig:seed_bench_overview}
\end{figure}

\begin{figure*}
    % \centering
    \includegraphics[width=1.0\textwidth]{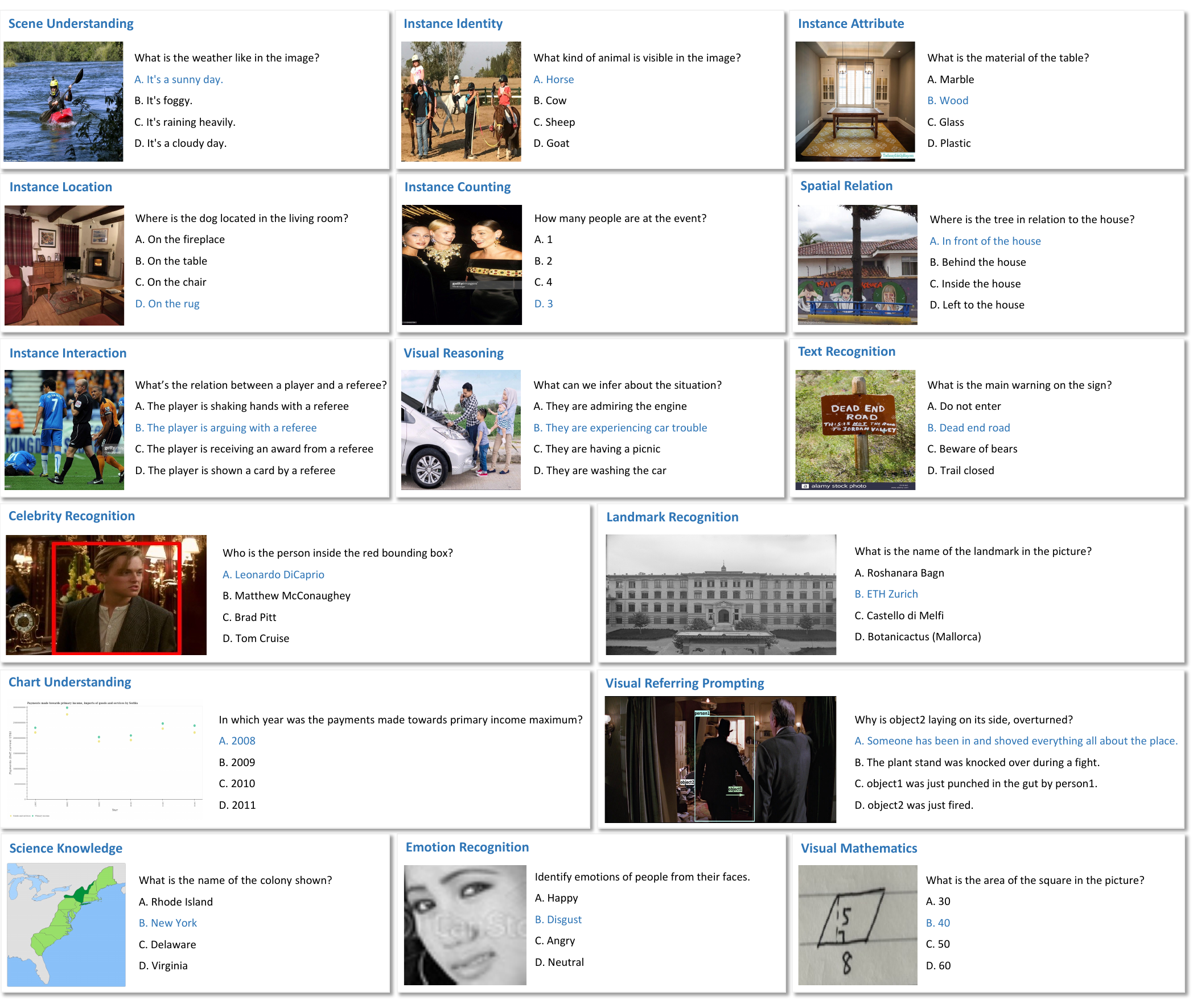}
    \caption{Data samples from a subset of evaluation dimensions in part-1 with single image as input, which encompasses capability $L_1$ in SEED-Bench-2.}
    \label{fig:part1_single}
        \vspace{-12pt}
\end{figure*}

\begin{figure*}
    \centering
    \includegraphics[width=0.87\textwidth]{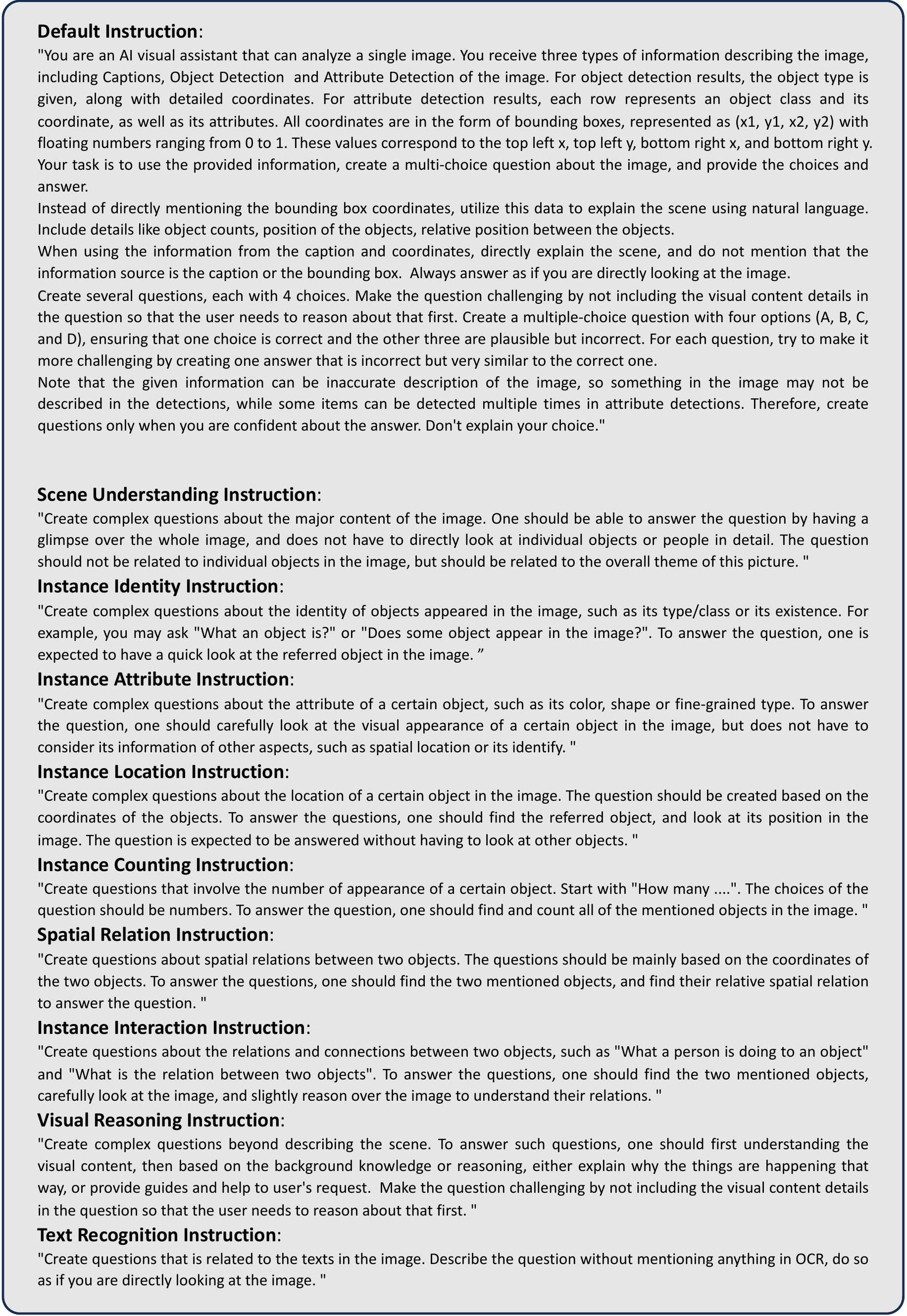}
    \caption{Prompts of generating multiple-choice questions for different evaluation dimensions.}
    \label{fig:prompt}
\end{figure*}

\begin{figure*}
    % \centering
    \includegraphics[width=1.0\textwidth]{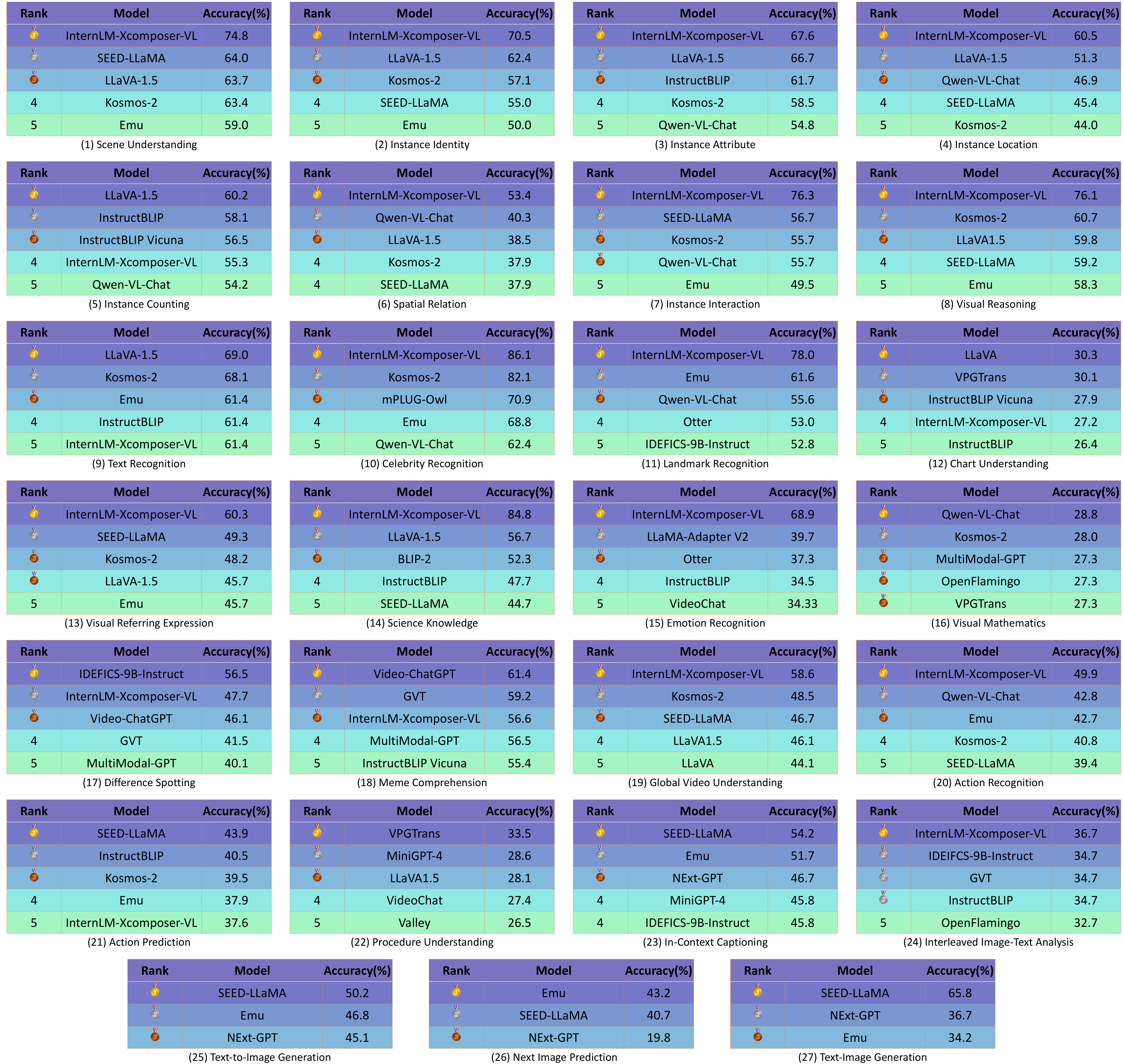}
    \caption{Each task leaderboard of SEED-Bench-2.}
    \label{fig:each_task_leader}
        \vspace{-12pt}
\end{figure*}

\begin{figure}
    % \centering
    \includegraphics[width=0.5\textwidth]{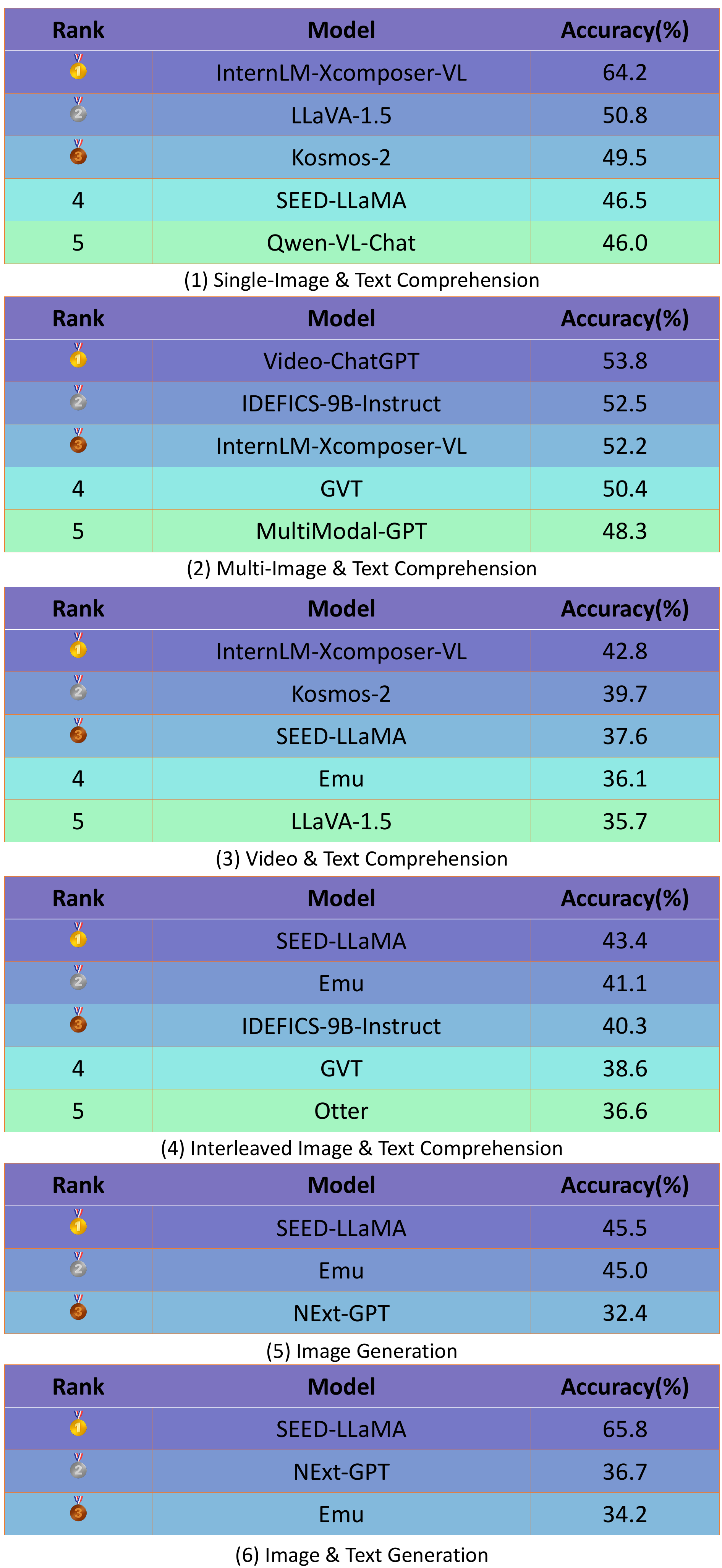}
    \caption{Subgroup task leaderboard of SEED-Bench-2.}
    \label{fig:subgroup_leader}
        \vspace{-12pt}
\end{figure}

\section{Evaluation Dimension}
\label{sec:evaluate_dimension}
To thoroughly evaluate the diverse capabilities of MLLMs, SEED-Bench-2 incorporates 27 assessment dimensions, encompassing Single-Image \& Text Comprehension, Multiple-Images \& Text Comprehension, Video \& Text Comprehension, Interleaved Image \& Text Comprehension, Image Generation, and Image \& Text Generation. The dimensions within Single-Image \& Text Comprehension, Multiple-Images \& Text Comprehension, and Video \& Text Comprehension are visually represented in Fig.~\ref{fig:seed_bench_overview}.

\noindent\textbf{Single-Image \& Text Comprehension.} The evaluation of single-image comprehension encompasses 16 dimensions, addressing global/object-level, recognition/reasoning, and various specialized domains.
\begin{itemize} 
    \item Scene Understanding: This dimension emphasizes global information in an image and necessitates a holistic understanding to answer questions about the overall scene. 
    \item Instance Identity: This dimension involves identifying specific instances in an image, including the existence or category of particular objects, evaluating a model's object recognition capabilities. 
    \item Instance Attribute: This dimension pertains to an instance's attributes, such as color, shape, or material, assessing a model's understanding of an object's visual appearance. 
    \item Instance Location: This dimension concerns the absolute position of a specified instance, requiring a model to accurately localize the object referred to in the question. 
    \item Instance Counting: This dimension necessitates that the model counts the number of specific objects in the image, understanding all objects and successfully counting the referred object's instances. 
    \item Spatial Relation: This dimension requires a model to ground two mentioned objects and recognize their relative spatial relation within the image. 
    \item Instance Interaction: This dimension involves recognizing the state relation or interaction relations between two humans or objects. 
    \item Visual Reasoning: This dimension evaluates a model's ability to reason based on visual information, necessitating a comprehensive understanding of the image and the application of commonsense knowledge to answer questions correctly. 
    \item Text Recognition: In this dimension, the model should answer questions about textual elements in the image. 
    \item Celebrity Recognition: This dimension focuses on identifying well-known public figures in images, evaluating a model's ability to recognize celebrity faces and names and understand their relevance in the given context. 
    \item Landmark Recognition: In this dimension, the model is required to recognize and identify famous landmarks or locations in the image, understanding visual features and contextual information associated with these landmarks. 
    \item Chart Understanding: This dimension requires the model to interpret and extract information from various chart types, such as line graphs, evaluating its ability to understand visual data representations and derive meaningful insights. 
    \item Visual Referring Expression: In this dimension, the model is required to answer relevant questions based on the visual content of the image, assessing its ability to understand the scene and engage in meaningful visual dialogue. 
    \item Science Knowledge: This dimension evaluates a model's ability to integrate multiple knowledge sources and apply commonsense reasoning to answer image-related questions, requiring an understanding of context, background information, and relationships between objects and events in the scene. 
    \item Emotion Recognition: This dimension focuses on recognizing and interpreting emotions expressed by human faces in images, evaluating the model's ability to understand facial expressions and associate them with corresponding emotional states. 
    \item Visual Mathematics: In this dimension, the model is required to solve mathematical problems or equations based on the visual content of the image, assessing its ability to understand and apply mathematical concepts and operations to real-world scenarios. 
\end{itemize}

\noindent\textbf{Multiple-Images \& Text Comprehension.} The evaluation of multiple-images comprehension comprises 2 dimensions: difference spotting and meme comprehension. These dimensions assess an MLLM's ability to extract information and discern differences from multiple images.
\begin{itemize} 
    \item Difference Spotting: In this dimension, the model is required to identify differences between two images, assessing its ability to recognize subtle variations in visual elements and understand the significance of these differences. 
    \item Meme Comprehension: This dimension requires the model to comprehend and interpret internet memes, which often involve humor, sarcasm, or cultural references. It evaluates the model's ability to recognize visual and textual meme elements and understand their intended meaning and context. 
\end{itemize}

\noindent\textbf{Video \& Text Comprehension.} For the evaluation of video comprehension, we propose 4 dimensions to assess an MLLM's ability to extract fine-grained information, temporal relationships, and reasoning through video content.

\begin{itemize} 
    \item Global Video Understanding: In this dimension, the model is required to answer questions from different aspects of a video's content, involving the understanding of key events, actions, and objects in the video, as well as recognizing their importance and relevance in the overall context of the video. 
    \item Action Recognition: This dimension requires the model to recognize actions shown in videos, evaluating its ability to capture temporal dynamics, physical motions, human actions, and dynamic interactions between objects. 
    \item Action Prediction: This dimension aims to predict future actions through preceding video segments, requiring an understanding of contextual information from videos and temporal reasoning. 
    \item Procedure Understanding: This dimension necessitates that the model captures key actions and performs temporal ordering on them, evaluating its ability for temporally fine-grained understanding and procedure reasoning. 
\end{itemize}

\noindent\textbf{Interleaved Image \& Text Comprehension.} For the evaluation of interleaved image-text data comprehension, we introduce 2 dimensions: in-context captioning and interleaved image-text analysis. These dimensions assess an MLLM's ability to extract information from arbitrary image-text data.
\begin{itemize} 
    \item In-Context Captioning: This dimension highlights a model's ability to learn and adapt its understanding based on the provided image context. It assesses the model's capacity to integrate new information, identify patterns, and generate predictions for the target image.
    \item Interleaved Image-Text Analysis: In this dimension, the model is required to process and understand data presented in an interleaved or mixed format, such as images combined with text. It assesses the model's ability to integrate multiple information modalities and derive meaningful insights from the combined data. 
\end{itemize}
    
\noindent\textbf{Image Generation.} To evaluate an MLLM's ability in image generation, we introduce two tasks: text-to-image generation and next image prediction. These tasks assess the MLLM's generation ability from text and multiple images.
\begin{itemize} 
    \item Text-to-Image Generation: This dimension evaluates a model's ability to generate realistic and visually coherent images based on a given prompt. It requires the model to understand visual elements, relationships, and composition rules necessary for creating a plausible image. 
    \item Next Image Prediction: In this dimension, the model is required to generate images that depict specific actions or events, such as a person running or a car driving. It assesses the model's ability to understand action dynamics and accurately represent them in a static visual format. 
\end{itemize}
    
\noindent\textbf{Image \& Text Generation.} To evaluate an MLLM's comprehensive ability in generation, we introduce the text-image creation task, which involves providing a question and requiring the MLLM to generate corresponding image and text as a description.
\begin{itemize} 
    \item Text-Image Creation: This dimension focuses on a model's ability to generate images with text. It evaluates the model's capacity to produce accurate text and visual content. 
\end{itemize}

\section{Data Source}
To create a benchmark with various evaluation dimensions, we need to collect data containing images with abundant visual information and videos with rich temporal dynamics, enabling us to construct diverse and challenging multiple-choice questions.

For dimensions 1-9, we utilize the CC3M~\cite{sharma2018conceptual_gcc} dataset with filtered samples to build questions for spatial understanding. Specifically, considering the noisy original captions of CC3M, we generate captions for each image with Tag2Text~\cite{huang2023tag2text}. We filter out images with no more than 5 nouns in their captions to ensure information richness in the remaining images for constructing questions. For limited data on text recognition, we use data from IC03~\cite{lucas2005ic03}, IC13~\cite{karatzas2013ic13}, IIIT5k~\cite{mishra2012IIIT5k}, and SVT~\cite{wang2011svt} datasets to enlarge this dimension.

For the celebrity recognition dimension, we use celebrity data from MME~\cite{fu2023mme} and MMBench~\cite{liu2023mmbench} to conduct this dimension. As celebrity recognition comprises 4-choice questions in MMBench and T/F questions in MME, we use GPT-4 to generate confusing options for MME data to construct 4-choice questions.

For the landmark recognition dimension, we use the Google landmark dataset v2~\cite{weyand2020google_landmark} train set as the data source and generate selections by randomly selecting other landmark names.

For the chart understanding dimension, we use the plotQA~\cite{methani2020plotqa} test set and generate selections using GPT-4 by inputting corresponding image captions.

For the visual referring expression dimension, we use the VCR~\cite{zellers2019vcr} valid dataset as the data source, and we use four methods to indicate the object in the picture: drawing a bounding box, drawing a circle, drawing a mask, and drawing an arrow.

For science knowledge, we use the scienceQA~\cite{lu2022scienceqa} test set, which contains image data for each question as the data source.

For emotion recognition, we use the fer2013~\cite{dumitru2013fer2013} test dataset as the image source and use the 6 emotions in the dataset as selections.

For visual mathematics, we use the math part of the MME~\cite{fu2023mme} dataset and generate some questions by human.

For difference spotting, we use the SD part of the MIMICIT~\cite{li2023mimic} dataset as the image source and generate selections using GPT-4.

For meme comprehension, we generate questions by human.

For global video understanding, we select the Charades~\cite{sigurdsson2016charades} test dataset as the video source, as the videos in the dataset contain rich information. For each video, we use tag2text~\cite{huang2023tag2text} to generate each second caption and grit~\cite{wu2022grit} to generate each 5-second dense caption containing each object's location. We then use GPT-4 to integrate captions and generate corresponding questions based on these captions. After generation, we use GPT-4 to filter out questions that can be answered using only a single frame.

For action recognition, and action prediction, we adopt Something-Something-v2 (SSV2)\cite{ssv2}, and Epic-kitchen 100~\cite{epickitchen100} datasets to build questions and let human annotators filter the questions. SSV2 is an action recognition dataset that includes 174 fine-grained categories of basic actions with everyday objects, and we adopt 1509 videos from its validation set. We also select 138 long videos from the Epic-kitchen 100 dataset with temporally annotated action labels. Moreover, videos and fine-grained action segmentation annotations in the Breakfast dataset~\cite{breakfast} are utilized for the procedure understanding task.

For in-context captioning, we use the ground-truth caption generated by instance attribute dimension and instance counting dimension. For each caption in the instance attribute, we use GPT-4 to classify.

For interleaved image-text analysis data, we generate questions by human.

For text-to-image generation, we firstly use GPT-4 to modify the target categories or attributes in prompt of CC-500~\cite{feng2022cc500} dataset and ABC-6k~\cite{feng2022cc500} dataset and form a four-choice question. We then use Stable-Diffusion-XL~\cite{podell2023sdxl} to generate each prompt and let human annotator to filter unqualified data.

For next image prediction dimension, we use  Epic-kitchen 100~\cite{epickitchen100} dataset and start-end frame in action prediction dimension to form this dimension.

For text-image creation, we generate questions by human.

\begin{table*}[]
    \centering
    \caption{Evaluation results of various MLLMs in 'Single-Image \& Text Comprehension' part of SEED-Bench-2. The best (second best) is in bold (underline).} \label{tab:single_image_task_performance}
    \vspace{3pt}
    % \vspace{-0.6em}
    {
    \resizebox{\textwidth}{!}{
    \begin{tabular}{cccccccccccccccccc}
         \toprule
         \multirow{1}{*}{Model} & \multirow{1}{*}{Language Model}& \makecell{Scene \\ Understanding} & \makecell{Instance \\ Identity} & \makecell{Instance \\ Attribute} & \makecell{Instance \\ Location} & \makecell{Instance \\ Counting} & \makecell{Spatial \\ Relation} & \makecell{Instance \\ Interaction} & \makecell{Visual \\ Reasoning} & \makecell{Text \\ Recognition} & \makecell{Celebrity \\ Recognition} & \makecell{Landmark \\ Recognition} & \makecell{Chart \\ Understanding} & \makecell{Visual \\ Referring \\ Expression} & \makecell{Science \\ Knowledge} &\makecell{Emotion \\ Recognition} & \makecell{Visual \\ Mathematics} \\
         \midrule
         BLIP-2~\cite{li2023blip2} &Flan-T5-XL &58.5 &48.6 &49.0 &39.1 & 43.4 &36.2 &48.5 &52.9 &60.7 &51.8 &51.4 &19.2 &43.2 &52.4 &29.3 &22.0\\
         InstructBLIP~\cite{dai2023instructblip} &Flan-T5-XL &58.9 &49.7 &61.7 &35.1 &58.1 &34.9 &47.4 &55.9 &61.4 &48.5 &45.4 &26.4 &41.7 &47.7 &34.5 &21.2\\
         InstructBLIP Vicuna~\cite{dai2023instructblip} &Vicuna-7B &53.6 &43.9 &49.0 &37.8 &\underline{56.5} &35.8 &43.3 &56.2 &57.2 &60.3 &44.4 &27.9 &39.2 &39.4 &23.0 &26.5\\
         LLaVA~\cite{liu2023visual_llava} &LLaMA-7B &53.8 &47.5 &38.3 &34.2 &42.0 &34.7 &40.2 &52.9 &46.4 &51.8 &45.6 &\bf{30.3} &40.2 &37.6 &34.3 &20.5\\
         MiniGPT-4~\cite{zhu2023minigpt4} &Vicuna-7B &56.3 &49.2 &45.8 &37.9 &45.3 &32.6 &47.4 &57.1 &41.8 &55.2 &45.2 &20.2 &41.2 &43.3 &24.2 &25.0\\
         VPGTrans~\cite{2023vpgtrans} &LLaMA-7B &46.9 &38.6 &33.6 &35.6 &27.5 &34.4 &33.0 &50.8 &47.6 &52.4 &38.2 &\underline{30.1} &34.7 &36.1 &31.5 &27.3\\
         MultiModal-GPT~\cite{gong2023multimodalgpt} &Vicuna-7B &46.9 &42.5 &32.0 &32.3 &27.7 &29.7 &29.9 &48.3 &35.2 &60.9 &50.4 &24.2 &42.2 &37.6 &32.1 &27.3\\
         Otter~\cite{li2023otter} &LLaMA-7B &45.9 &39.7 &31.9 &31.6 &26.4 &32.0 &33.0 &49.2 &39.3 &59.7 &53.0 &23.6 &41.2 &36.1 &37.3 &22.0\\
         OpenFlamingo~\cite{openflamingo} &LLaMA-7B &46.7 &42.3 &31.7 &33.4 &27.4 &29.8 &29.9 &47.7 &35.6 &60.3 &49.8 &24.2 &42.2 &39.0 &32.1 &27.3\\
         LLaMA-Adapter V2~\cite{gao2023llamaadapterv2} &LLaMA-7B &45.2 &38.5 &29.3 &33.0 &29.7 &35.5 &39.2 &52.0 &48.7 &58.5 &46.4 &24.2 &41.2 &40.1 &\underline{39.7} &23.5\\
         GVT~\cite{wang2023gvt} &Vicuna-7B &41.7 &35.5 &31.8 &29.5 &36.2 &32.0 &32.0 &51.1 &35.2 &39.4 &36.4 &25.0 &36.2 &31.1 &20.6 &22.7\\
         mPLUG-Owl~\cite{ye2023mplugowl} &LLaMA-7B &49.7 &45.3 &32.5 &36.7 &27.3 &32.7 &44.3 &54.7 &49.2 &70.9 &49.6 &23.2 &44.2 &44.0 &32.5 &23.5\\ 
         Kosmos-2~\cite{peng2023kosmos} & Decoder only 1.3B &63.4 &57.1 &58.5 &44.0 &41.4 &37.9 &{55.7} &\underline{60.7} &\underline{68.1} &\underline{82.1} &51.4 &21.2 &{48.2} &43.7 &30.7 &\underline{28.0} \\
         Qwen-VL-Chat~\cite{bai2023qwen} & Qwen-7B &56.5 &47.6 &54.8 &46.9 &54.2 &\underline{40.3} &55.7 &55.0 &47.4 &62.4 &55.6 &25.2 &43.7 &41.2 &20.6 &\bf{28.8}\\
         LLaVA-1.5~\cite{liu2023llava1.5} &Vicuna-7B  &{63.7} &\underline{62.4} &\underline{66.7} &\underline{51.3} &\bf{60.2} &38.5 &47.4 &59.8 &\bf{69.0} &60.6 &49.8 &25.0 &45.7 &\underline{56.7} &31.1 &24.2\\
         IDEFICS-9B-Instruct~\cite{laurencon2023obelics} & LLaMA-7B &48.2 &38.2 &37.8 &32.9 &29.0 &32.4 &37.1 &54.1 &45.5 &52.4 &52.8 &22.6 &42.7 &33.2 &26.6 &21.2\\
         InternLM-Xcomposer-VL~\cite{zhang2023internlm} & InternLM-7B &\bf{74.8} &\bf{70.5} &\bf{67.6} &\bf{60.5} &55.3 &\bf{53.4} &\bf{76.3} &\bf{76.1} &61.4 &\bf{86.1} &\bf{78.0} &27.2 &\bf{60.3} &\bf{84.8} &\bf{68.9} &25.8\\
         VideoChat~\cite{li2023videochat} &Vicuna-7B &44.3 &40.7 &32.2 &36.9 &32.9 &32.6 &42.3 &51.1 &45.8 &35.2 &46.8 &20.6 &43.2 &39.4 &34.3 &19.7\\
         Video-ChatGPT~\cite{maaz2023videochatgpt} &LLaMA-7B &44.1 &37.0 &35.8 &30.7 &44.2 &31.1 &29.9 &49.9 &39.8 &49.7 &40.6 &22.0 &33.2 &37.2 &22.4 &25.0\\
         Valley~\cite{luo2023valley} &LLaMA-13B &45.3 &36.4 &33.7 &30.6 &27.1 &31.5 &35.1 &52.0 &35.2 &44.9 &43.4 &23.8 &33.2 &37.2 &26.0 &22.7\\
         Emu~\cite{sun2023emu} & LLaMA-13B &59.0 &50.0 &43.7 &37.1 &44.3 &33.6 &49.5 &58.3 &61.4 &68.8 &\underline{61.6} &19.0 &45.7 &41.5 &24.2 &26.4\\
         NExt-GPT~\cite{wu2023nextgpt} &Vicuna-7B &36.4 &35.1 &25.6 &29.9 &36.1 &30.9 &39.2 &41.7 &31.0 &30.9 &27.4 &21.2 &34.2 &31.8 &24.4 &17.4\\
         SEED-LLaMA~\cite{ge2023making} & LLaMA2-Chat-13B & \underline{64.0} & 55.0 & 51.3 & 45.4 & 43.3 & 37.9 & \underline{56.7} & 59.2 & 57.0 & 55.5 & 52.8 & 18.8 & \underline{49.3} & 44.8 & 28.8 & 24.4 \\
         \bottomrule
    \end{tabular}
    }
   }
\end{table*}

\begin{table*}[]
    \centering
    \caption{Evaluation results of various MLLMs in 'Multi-Images \& Text Comprehension' part, 'Video \& Text Comprehension' part, 'Interleaved Image \& Text Comprehension' part, 'Image Generation' part, 'Image \& Text Generation' part of SEED-Bench-2. The best (second best) is in bold (underline).}\label{tab:other_task_performance}
    \vspace{3pt}
    % \vspace{-0.6em}
    {
    \resizebox{\textwidth}{!}{
    \begin{tabular}{ccccccccccccc}
         \toprule
         \multirow{6}{*}{Model} & \multirow{6}{*}{Language Model} &\multicolumn{6}{c}{part 1} &\multicolumn{2}{c}{part 2} &\multicolumn{3}{c}{part 3}\\
         \cmidrule(lr){3-8}
         \cmidrule(lr){9-10}
         \cmidrule(lr){11-13}
         && \multicolumn{2}{c}{\makecell{Multi-Images \&\\ Text Comprehension}} & \multicolumn{4}{c}{\makecell{Video \& \\Text Comprehension}} & \multicolumn{2}{c}{\makecell{Interleaved Image \&\\ Text Comprehension}} & \multicolumn{2}{c}{\makecell{Image \\Generation}} & \makecell{Image \& Text \\Generation} \\
         \cmidrule(lr){3-4}
         \cmidrule(lr){5-8}
         \cmidrule(lr){9-10}
         \cmidrule(lr){11-12}
         \cmidrule(lr){13-13}
         && \makecell{Difference \\ Spotting} & \makecell{Meme \\ Comprehension} & \makecell{Global Video \\ Understanding} & \makecell{Action \\ Recognition} & \makecell{Action \\ Prediction} & \makecell{Procedure \\ Understanding} & \makecell{In-Context \\ Captioning} & \makecell{Interleaved \\ Image-Text \\ Analysis} & \makecell{Text-to-Image \\ Generation} & \makecell{Next Image \\ Prediction} & \makecell{Text-Image \\ Creation} \\
         \midrule
         BLIP-2~\cite{li2023blip2} &Flan-T5-XL &17.8 &38.6 &42.5 &37.7 &36.2 &22.9 &40.0 &30.6 &- &- &-\\
         InstructBLIP~\cite{dai2023instructblip} &Flan-T5-XL &22.8 &35.2 &41.5 &36.1 &\underline{40.5} &24.5 &36.7 &\underline{34.7} &- &- &-\\
         InstructBLIP Vicuna~\cite{dai2023instructblip} &Vicuna-7B &36.5 &55.4 &40.4 &38.6 &31.2 &15.6 &26.7 &32.7 &- &- &-\\
         LLaVA~\cite{liu2023visual_llava} &LLaMA-7B &27.0 &50.0 &44.1 &36.2 &25.1 &18.6 &40.0 &20.4 &- &- &-\\
         MiniGPT-4~\cite{zhu2023minigpt4} &Vicuna-7B &19.0 &46.7 &39.0 &38.7 &27.4 &28.6 &\underline{45.8} &22.5 &- &- &-\\
         VPGTrans~\cite{2023vpgtrans} &LLaMA-7B &24.6 &44.0 &37.8 &38.2 &20.9 &\bf{33.5} &19.2 &28.6 &- &- &-\\
         MultiModal-GPT~\cite{gong2023multimodalgpt} &Vicuna-7B &40.1 &56.5 &37.6 &38.7 &25.3 &24.4 &39.2 &30.6 &- &- &-\\
         Otter~\cite{li2023otter} &LLaMA-7B &27.4 &46.7 &36.6 &37.9 &26.0 &24.8 &42.5 &30.6 &- &- &-\\
         OpenFlamingo~\cite{openflamingo} &LLaMA-7B &39.9 &54.9 &37.6 &38.4 &25.2 &24.1 &38.3 &32.7 &- &- &-\\
         LLaMA-Adapter V2~\cite{gao2023llamaadapterv2} &LLaMA-7B &29.1 &52.2 &41.9 &38.2 &18.8 &20.3 &- &- &- &- &-\\
         GVT~\cite{wang2023gvt} &Vicuna-7B &41.5 &\underline{59.2} &40.4 &29.7 &26.3 &24.1 &42.5 &34.7 &- &- &-\\
         mPLUG-Owl~\cite{ye2023mplugowl} &LLaMA-7B &33.5 &54.9 &42.0 &37.8 &18.3 &19.3 &29.2 &28.6 &- &- &-\\ 
         Kosmos-2~\cite{peng2023kosmos} & Decoder only 1.3B &25.2 &42.8 &\underline{48.5} &40.8 &{39.5} &\underline{30.0} &24.2 &22.5 &- &- &-\\
         Qwen-VL-Chat~\cite{bai2023qwen} & Qwen-7B &34.3 &47.2 &39.7 &\underline{42.8} &29.6 &19.1 &42.5 &28.6 &- &- &-\\
         LLaVA-1.5~\cite{liu2023llava1.5} &Vicuna-7B &35.7 &50.3 &46.1 &39.4 &29.4 &28.1 &39.2 &22.5 &- &- &-\\
         IDEFICS-9B-Instruct~\cite{laurencon2023obelics} & LLaMA-7B &\bf{56.5} &48.4 &42.7 &38.6 &23.6 &20.5 &{45.8} &\underline{34.7} &- &- &-\\
         InternLM-Xcomposer-VL~\cite{zhang2023internlm} & InternLM-7B &\underline{47.7} &{56.6} &\bf{58.6} &\bf{49.9} &37.6 &24.9 &27.5 &\bf{36.7} &- &- &-\\
         VideoChat~\cite{li2023videochat} &Vicuna-7B &30.3 &51.6 &41.5 &34.0 &30.6 &27.4 &40.0 &30.6 &- &- &-\\
         Video-ChatGPT~\cite{maaz2023videochatgpt} &LLaMA-7B &46.1 &\bf{61.4} &42.6 &32.2 &27.0 &19.0 &37.5 &24.5 &- &- &-\\
         Valley~\cite{luo2023valley} &LLaMA-13B &37.1 &52.2 &31.5 &32.1 &21.9 &26.5 &35.8 &28.6 &- &- &-\\
         Emu~\cite{sun2023emu} & LLaMA-13B &29.3 &37.1 &41.9 &42.7 &37.9 &21.8 &51.7 &30.6 &\underline{46.8} &\bf{43.2} &{34.2}\\
         NExt-GPT~\cite{wu2023nextgpt} &Vicuna-7B &24.2 &39.0 &35.5 &33.8 &25.6 &24.5 &46.7 &24.5 &{45.1} &{19.8} &\underline{36.7} \\
         SEED-LLaMA~\cite{ge2023making} & LLaMA2-Chat-13B & 29.5 & 41.5 & 46.7 & 39.4 & \bf{43.9} & 20.3 & \bf{54.2} & 32.7 &\bf{50.2} &\underline{40.7} &\bf{65.8}\\
         \bottomrule
    \end{tabular}
    }
   }
\end{table*}

\begin{figure}
    % \centering
    \includegraphics[width=0.5\textwidth]{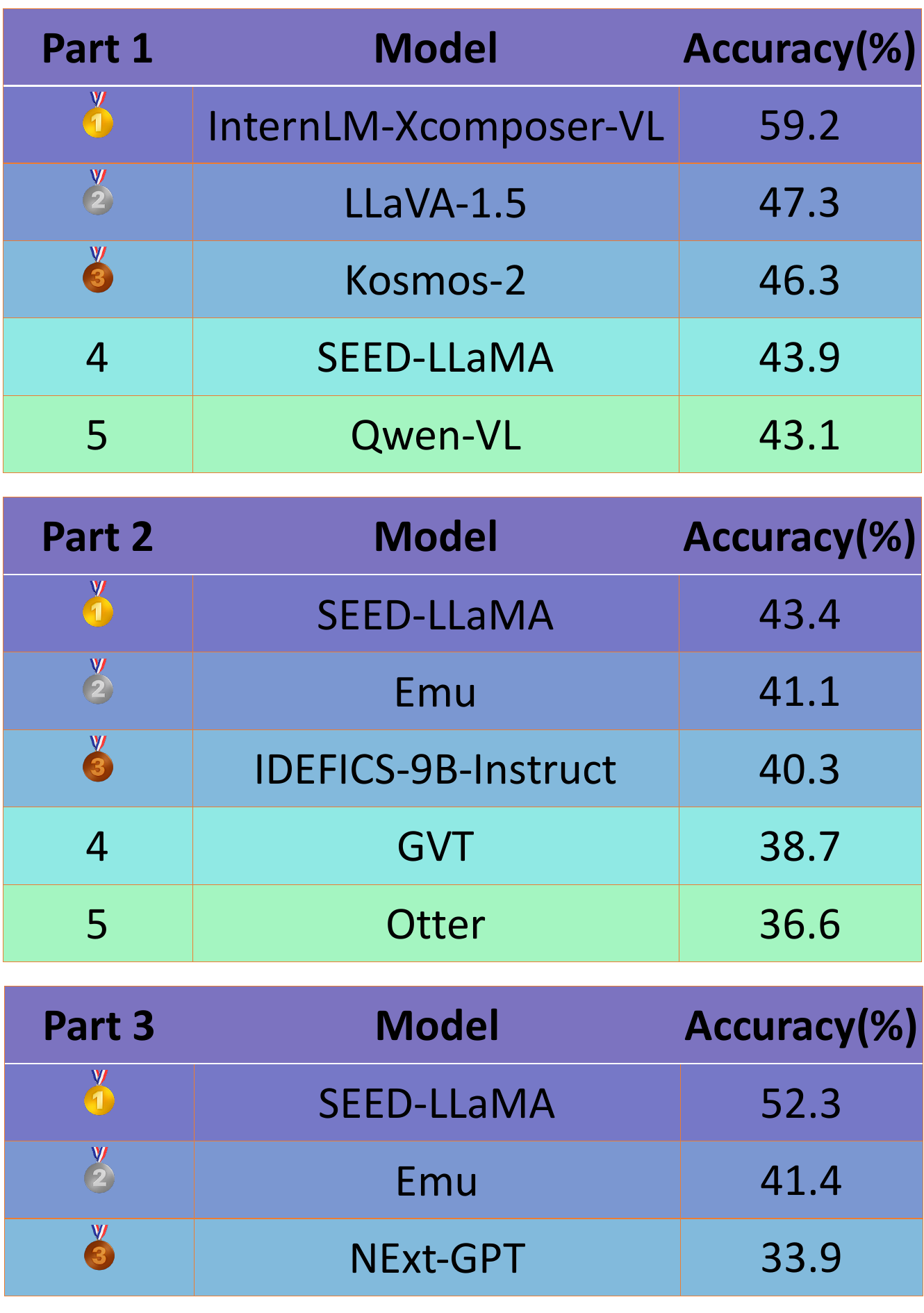}
    \caption{Part leaderboard of SEED-Bench-2.}
    \label{fig:part_leader}
        \vspace{-12pt}
\end{figure}

\section{Automatic Pipeline}
In this section, we provide a detailed discussion of automatic pipeline for constructing multiple-choice questions for dimensions 1-9.

\noindent\textbf{Visual Information Extraction.}
% Rich information in the image should be extracted into text formats, so that large-scale pretrained models can sufficiently understand the contents in the image to create meaningful and challenging questions.
For constructing questions related to spatial understanding, we interpret the rich information in each image with texts using multiple pretrained models, so that ChatGPT/GPT-4 can understand the image and create questions accordingly. 
% For constructing questions related to temporal understanding, considering that extracting reliable temporal information from videos (especially fine-grained actions and long-term temporal context) is extremely difficult given existing foundation models, we utilize the ground-truth annotations of video datasets. We will explore how to generate questions based on automatically extracted video information in the future.
The extraction of visual information for images includes the following parts:

\begin{itemize}
\item \noindent\textbf{Image Captions.} 
Image captions contain the overall description of an image. 
% In order to generate captions that describe the image from different angles, we utilized multiple image captioning model.
We employ BLIP2~\cite{li2022blip} and Tag2Text~\cite{huang2023tag2text} to create captions for each image.
The former creates captions for the whole image while the latter generates captions based on descriptions of each instance.
% We first use BLIP2~\cite{li2023blip2}, which is widely used to generate image captions in various applications.
% The other model we utilized is Tag2Text~\cite{huang2023tag2text}, which creates image captions in a bottom-up manner.
The two models complement each other to depict the image content within a single sentence.

\item \noindent\textbf{Instance Descriptions.}
Besides captions which may ignore specific details in the image, we also extract visual information from images using instance-level descriptions, including object detection, attribute detection, and dense captions.
Specifically, we use SAM~\cite{kirillov2023sam} to segment each instance in the image and obtain their bounding boxes according to the segmentation results.
The object labels are obtained using Tag2Text~\cite{huang2023tag2text}.
Besides, we also utilize attribute detector~\cite{zhang2021vinvl} to obtain the attributes of each instance in the image.
Finally, we employ GRiT~\cite{wu2022grit} to generate dense captions, which describe each detected instance in the image with a short sentence.
These instance-level descriptions are complementary to the image captions, further enriching the visual information of each image. 

\item \noindent\textbf{Textual Elements.}
Besides objects, the texts in the image also contain important information describing the image. 
We employ PaddleOCR~\cite{paddleocr} for detecting textual elements.
\end{itemize}

\noindent\textbf{Question-Answer Generation.} After extracting visual information from the image, we task ChatGPT/GPT-4 with generating multiple-choice questions based on the extracted information or video annotations.  
For each of the spatial understanding evaluation, we carefully design prompts and ask ChatGPT/GPT-4 to create multiple choice questions with four candidate options based on the extracted visual information.
We create questions with ChatGPT for all evaluation dimensions, except for the reasoning dimension, where we use GPT-4~\cite{openai2023gpt4} due to its exceptional reasoning capability. 
For each question, we ask ChatGPT/GPT-4 to create four choices with one correct option and three distractors. We try to make the multiple-choice questions challenging by encouraging the three wrong choices to be similar to the correct one. The detailed prompts of generating multiple-choice questions for different evaluation dimensions are listed in Fig.~\ref{fig:prompt}.
% For generating questions related to temporal understanding, we utilize the ground-truth annotations of selected videos as the answer of multi-choice questions and employ ChatGPT to generate three distractors.

%temporal understanding tasks, we employ the ground-truth annotations of selected videos as the answer of multi-choice questions and generate 3 wrong choices with ChatGPT based on given labels and background information. Additionally, for each task, the question of each QA pair is randomly selected from multiple templates generated by ChatGPT, given the setting of each task. 

\noindent\textbf{Automatic Filtering.}
Our benchmark aims at evaluating the multimodal vision-language understanding capability of MLLMs. 
However, we observe that some generated questions can be correctly answered by LLMs without seeing the image.
We argue that such questions are not helpful to evaluate the visual comprehension capability of MLLMs.
To this end, we feed the generated questions (without image) into three powerful LLMs, including Vicuna-7B~\cite{vicuna}, Flan-T5-XXL~\cite{chung2022scaling_flant5} and LLaMA-7B~\cite{touvron2023llama} and ask them to answer the questions.
We empirically found that $5.52\%$ of the generated questions can be correctly answered by all of the three LLMs.
We filter out these questions from our benchmark.

\noindent\textbf{Human Annotation.}
To ensure the accuracy and objectiveness of SEED-Bench-2, we further employ human annotators to verify the generated question/answer pairs.  Human annotators are asked to choose the correct answer for each multiple-choice question and categorize each question into one of the evaluation dimension. If one question can not be answered based on the visual input or does not have any correct choice or has multiple correct choices, it will be discarded by human annotators.

\section{Evaluation Results}
Detailed evaluation result for 23 models in 27 tasks is presented in Tab.~\ref{tab:single_image_task_performance} and Tab.~\ref{tab:other_task_performance}. In these tables, the best and second-best performances for each task are indicated in bold and underlined, respectively.

Additionally, the leaderboards for each task, sub-part, and part are displayed in  Fig.~\ref{fig:each_task_leader}, Fig.~\ref{fig:subgroup_leader}, and Fig.~\ref{fig:part_leader}.